\documentclass[onecolumn,10pt]{article}
\usepackage{palatino} 
\usepackage[T1]{fontenc}
\usepackage{xcolor,fullpage}
\usepackage{graphicx}
\usepackage{fancyhdr}
\usepackage{amsmath}
\usepackage{amssymb}
\usepackage{textcomp}
\usepackage[utf8]{inputenc}
\usepackage{subcaption}	
\usepackage{hyperref}
\usepackage{setspace}
\usepackage{cancel}
\usepackage{mathtools}
\usepackage[font=footnotesize]{caption}
\usepackage{rotating}
\usepackage{adjustbox}
\usepackage{multirow}
\usepackage{tabularx}
\usepackage{lipsum}
\usepackage{booktabs}
\usepackage{subfiles}
\usepackage{titling}
\usepackage{placeins}
\usepackage{authblk}
\usepackage{comment}
\usepackage{wrapfig}
\usepackage{chngcntr}
\usepackage{algorithm}
\usepackage{algpseudocode}

\usepackage[square,sort,comma,numbers]{natbib}
\usepackage[acronym]{glossaries}

\newacronym[plural=FEAs]{fea}{FEA}{Finite Element Analysis}
\newacronym{sindy}{SINDy}{Sparse Identification of Nonlinear Dynamics}
\newacronym{stlsq}{STLS}{Sequential Threshold Least Squares}
\newacronym[plural=ODEs]{ode}{ODE}{Ordinary Differential Equation}
\newacronym[plural=PDEs]{pde}{PDE}{Partial Differential Equation}
\newacronym{sr3}{SR3}{Sparse Relaxed Regularized Regression}
\newacronym{doe}{DoE}{Design of Experiments}
\newacronym{lhs}{LHS}{Latin Hypercube Sampling}
\newacronym{uq}{UQ}{Uncertainty Quantification}
\newacronym{si}{SI}{System Identification}
\newacronym{al}{AL}{Active Learning}
\newacronym{es}{E-SINDy}{Ensemble SINDy}
\newacronym{fft}{fft}{Fast Fourier Transform}

\makeglossaries

\newcommand{\keywords}[1]{\par\vspace{0.5em}{\centering \textit{\textbf{Keywords: }}#1\par}}

\newcolumntype{L}[1]{>{\raggedright\let\newline\\\arraybackslash\hspace{0pt}}m{#1}}
\newcolumntype{C}[1]{>{\centering\let\newline\\\arraybackslash\hspace{0pt}}m{#1}}
\newcolumntype{R}[1]{>{\raggedleft\let\newline\\\arraybackslash\hspace{0pt}}m{#1}}

\title{\textbf{\LARGE{How Low Can You Go? Active Learning for Sparse Model Discovery in the Ultra-Low-Data Limit}}\vspace{-.25in}}
\author[1,2]{Ana Larrañaga$^*$}
\author[3]{Urban Fasel}
\author[1,2]{Steven L. Brunton$^*$}

\affil[1]{\small Department of Mechanical Engineering, University of Washington, Seattle, WA 98195, United States}
\affil[2]{\small NSF AI Institute in Dynamic Systems, University of Washington, Seattle, WA 98195, United States}
\affil[3]{\small Department of Aeronautics, Imperial College London, SW7 2AZ, United Kingdom}
\affil[$*$]{{\footnotesize corresponding author: alarra@uw.edu, sbrunton@uw.edu}}

\begin{document}

\date{}
\maketitle
\vspace{-.75in}
\begin{abstract}
\normalsize
Identifying the governing equations of complex dynamical systems remains a fundamental challenge across science and engineering. While early approaches relied on empirical data and heuristics, modern data-driven methods offer greater flexibility and fewer assumptions. However, data acquisition in real-world settings is often expensive. This work addresses this challenge by introducing an active learning strategy for dynamics discovery in the ultra-low data limit. Rather than sampling randomly, our method iteratively prioritizes regions that are most informative for model identification. This approach builds on \gls{sindy}, and utilizes an ensemble extension, E-SINDy, to estimate epistemic uncertainty and guide the sampling for both ordinary and partial differential equations (ODEs/PDEs). For ODEs, an exhaustive analysis is conducted on the Lorenz system across varying data budgets and noise levels. For PDEs, two systems with contrasting dynamical characteristics are examined: the Burgers' equation, where a sharp shock front creates a distinction between informative and uninformative regions, and the Kuramoto--Sivashinsky equation, which presents a more spatially complex sampling landscape. Across all scenarios, the proposed method accurately identifies the governing dynamics with significantly fewer data samples than random sampling.
\end{abstract}

\vspace{0.3cm}

A central question in modeling complex nonlinear systems is how much data is needed to accurately capture the underlying dynamics~\cite{Brunton_Kutz_2019, shumaylov2025}. In many applications, data sampling strongly affects how well the system’s underlying patterns are uncovered~\cite{brunton2016}, especially when the data is noisy~\cite{Angluin1988, Madu2022}, which is generally the case for real-world applications. In chaotic systems, where small changes in initial conditions can lead to very different outcomes, choosing the right sampling strategy becomes even more important~\cite{Drgona2025, champion2019, Wagenmaker2024}. This is particularly true in data-driven models, where performance depends heavily on how representative and varied the data is. A good sampling approach must also balance exploring different system behaviors and improving knowledge where the system is already partially understood~\cite{zhao22a}. These questions become even more relevant when data collection is expensive in terms of time or computing power, as is often the case in industrial engineering applications, such as material science~\cite{Xu2023}, or when modeling extremely low-frequency events (e.g. structural failures, space weather, or pandemic surges)~\cite{Pickering2022, camporeale2023}.

In many fields, researchers face a common challenge: \textit{how can we learn the system's governing equations with the minimal amount of information?} Recent advances in machine learning have made it possible to uncover latent structures in high-dimensional datasets~\cite{Willard2022, Willcox2023, Dellapia2024}, that encode complex patterns often inaccessible to classical techniques. Acquiring high-quality data is costly: high-fidelity simulations demand substantial computational resources, and experimental campaigns are also time- and resource-intensive. When data collection is limited or expensive, efficiency becomes essential~\cite{musekamp2024, do2019}. Hence, we want to learn the best model as quickly as possible and adjust it as we gather more information~\cite{Schaeffer2020}, without assuming we can simply collect large amounts of information. And even though the objective is clear, the problem itself is a challenging optimization task, that fundamentally can be seen as an inverse problem, ill-posed, that inherently contains uncertainty from the sparse and noisy training data. Interestingly, recent work has demonstrated that this uncertainty can serve as a feature that characterizes the system in a more meaningful way, rather than being treated as a flaw~\cite{seung1992, Fasel2022}.

\gls{doe} is not a new idea. In 1937, Fisher established the foundation of modern experimental design by introducing the principles of randomization, replication, and local control~\cite{fisher1937design}. Combined with the goal of reducing model uncertainty, several optimality criteria were later introduced to maximize the information from the Fisher information matrix and guide sampling~\cite{atkinson1992}. A-, E-, D- optimality are just a few examples of these criteria that mainly focus on shaping the eigen-structure of said matrix~\cite{kiefer1959, Atkinson2007}, which can help reduce issues related to poor conditioning~\cite{Brunton_Kutz_2019}. However, classical \gls{doe} relies on planning experiments in advance and assuming a model structure, which becomes limiting when data is scarce and the underlying system is poorly understood. This motivated adaptive, sequential approaches that iteratively decide where to sample next based on what has been learned---a methodology that emerged in both statistics (as sequential \gls{doe}~\cite{box1992sequential}) and machine learning (as active learning~\cite{Xu2024, settles2012}).

\gls{al} focuses on selecting informative and diverse samples to reduce the amount of training data needed without loss of accuracy. A typical \gls{al} loop consists of (1) selecting a sample, (2) updating the model, and (3) evaluating performance, which supports a fast and adaptable training in online and control scenarios. For it to be successful, samples need to be (1) informative, (2) representative, and (3) diverse~\cite{musekamp2024, do2019}, using either uncertainty estimates or a feature-based approach~\cite{holzmuller2023} for their selection~\cite{Fu2013}. This sequential procedure exploits accumulated information and has been shown to be effective~\cite{adcock2023}, and mathematically reliable for \gls{si}~\cite{mania2022, shields2023} and the discovery of dynamical systems~\cite{Wagenmaker2024, Elend2023}. Among data-driven methods for identifying governing equations, \gls{sindy} is widely and successfully used in many fields~\cite{Delabays2025, Horrocks2020}. This algorithm uses sparse regression to find a parsimonious representation of systems dynamics~\cite{brunton2016}, being specially useful for control~\cite{Kaiser2018}, noisy data~\cite{Fasel2022}, and working in the low data limit~\cite{champion2019, Zolman2025}. 

For the sample selection based on uncertainty, the actual model prediction is used as the estimate, being directly calculated as part of the output (using Bayesian regression methods~\cite{DiFiore2024}, including Gaussian Process Regression~\cite{Fan2019}), obtained from the disagreement (variance) of an ensemble of models~\cite{krogh1994, Fasel2022}, or using novel statistical approaches that can quantify the uncertainty of predictions made by any algorithm~\cite{Vovk2005, Fasel2025}. The former alternatives, Bayesian approaches, are often impractical in real-world settings due to their high training cost and poor scalability with larger datasets. In contrast, methods like \gls{sindy} are far more efficient, which has contributed to the increasing interest in ensemble-based alternatives. Nonetheless, researchers have investigated alternatives that combine \gls{sindy} with Gaussian approximations~\cite{fung2025}.

In this context, we adopt \gls{es} as our surrogate model because it allows us to obtain reliable uncertainty estimates using an ensemble-based strategy~\cite{Fasel2022}. Importantly, this choice is not restrictive, since the overall approach is also compatible with other fast and accurate surrogate architectures. Our focus in this work is the ultra-low-data regime, where we aim to quantify the minimum data requirements for both \gls{ode} and \gls{pde} settings. We also introduce a practical convergence criterion for combining active learning with \gls{es}, which, to our knowledge, has not yet been examined in depth in the literature. By applying our ensemble-based \gls{sindy} method to a set of \gls{ode} and \gls{pde} benchmarks and systematically varying noise and data levels, we identify when active learning exceeds the performance of random sampling and establish its practical value in low-data settings.

\vspace{-0.6cm}

\section{Background}
\vspace{-0.4cm}
\label{background}
Here, we provide an overview of sparse dictionary learning, focusing on \gls{sindy} and its extensions. We present how active learning can be incorporated into the search for the governing equations of a dynamical system. Special attention is given to the challenges and complexities associated with selecting an appropriate sampling criteria.
\vspace{-0.5cm}
\subsection{Sparse dictionary learning for dynamics discovery}
\label{sindy}
\vspace{-0.4cm}
Denoting the state of a system at time \textit{t} by $\mathbf{x}(t)\in \mathbb{R}^{n}$, we consider the dynamical system $\dot{\mathbf{x}}(t) = \mathbf{f}(\mathbf{x}(t))$, where $\mathbf{f} : \mathbb{R}^{n} \to \mathbb{R}^{n}$ represents governing equations to be learned. To do so, the state $\mathbf{x}(t)$ is sampled at multiple time instants $t_{1},...,t_{m}$ and the corresponding derivatives $\dot{\mathbf{x}}(t)$ are either measured directly or approximated numerically. This results in a matrix of sampled states $\mathbf{X}\in \mathbb{R}^{m \times n}$ and a corresponding matrix of time derivatives $\dot{\mathbf{X}}\in \mathbb{R}^{m \times n}$, where each row contains $\mathbf{x}(t_{i})^\top$ and $\dot{\mathbf{x}}(t_{i})^\top$, respectively. 

\gls{sindy} is a specific case of sparse dictionary learning that assumes the underlying dynamics of a physical system can be expressed as a sparse linear combination of candidate linear or non-linear functions. Given a library of candidate functions $\Theta(\mathbf{X}) \in \mathbb{R}^{m \times p}$, which may include constants, polynomials, or trigonometric terms, we approximate the time derivatives as $\dot{\mathbf{X}} = \Theta(\mathbf{X})\Xi$. The matrix $\Xi \in \mathbb{R}^{p \times n}$ contains the coefficients of the equations, and its columns $\xi \in \mathbb{R}^{p}$ are sparse because only a small subset of library functions is assumed to be active for each component of the state. Its objective is to find a parsimonious set of differential equations that describe observed time-series data. 

Building on this definition, we define the dynamical system as $\dot{\mathbf{x}}(t) = \mathbf{f}(\mathbf{x}(t))=\Xi^\top(\Theta(\mathbf{x}^\top))^\top $. The function $\mathbf{f}$ may describe a finite-dimensional \gls{ode} or a spatial discretization of a \gls{pde}. For an ODE, the library $\Theta(\mathbf{X})$ typically contains functions of the state only, for example:
\begin{equation}
    \Theta(\mathbf{X}) = \big[\,\mathbf{1},\ \mathbf{X},\ \mathbf{X}^{2},\ \mathbf{X}^{3},\ \sin(\mathbf{X}),\ \ldots \big],
\end{equation}
where expressions such as $\mathbf{X}^{2}$ or $\sin(\mathbf{X})$ are interpreted element-wise, meaning that each nonlinear function is applied independently to every component of the state. For a PDE, where the dynamics depend on spatial derivatives, the library must also include candidate terms involving spatial derivatives of the field variable $u$, for example:
\begin{equation}
    \Theta(\mathbf{X}) = \big[\,\mathbf{1},\ u,\ u^{2},\ u_{x},\ u_{xx},\ u\,u_{x},\ u^{2}u_{xx},\ \ldots \big],
\end{equation}
with $u$, $u_{x}$, and $u_{xx}$ representing the discretized field and its first and second spatial derivatives. In this way, the structure of $\Theta$ reflects whether the system is governed solely by temporal dynamics, as in \glspl{ode}, or by spatiotemporal dynamics, as in \glspl{pde}.

Moreover, the coefficient matrix $\Xi$ contains the sparse coefficients that identify which terms in the library $\Theta(\mathbf{X})$ are active in the governing dynamics of $\mathbf{X}$. It is estimated as a regression problem with a sparsity-promoting regularization $\mathcal{R}$: 
\begin{equation}
    \hat{\Xi} = \arg\min_{\hat{\Xi}} \| \mathbf{\dot{\mathbf{X}}} - \Theta(\mathbf{X}) \hat{\Xi} \|_2^2 + \mathcal{R}(\hat{\Xi})
    \label{eq_reg}
\end{equation}
Such regularization can be performed in different ways, with the most common one being the classical sequential thresholded least squares (STLSQ) algorithm proposed in the original \gls{sindy} formulation~\cite{brunton2016}. STLSQ solves the least-squares problem first (first term in eq.\eqref{eq_reg}) and then sets to zero small coefficients based on a threshold parameter $\lambda$ that controls the sparsity of the expressions. Other approaches include sequential thresholded ridge regression (STRidge)~\cite{rudy2017} or sparse relaxed regularized regression (SR3)~\cite{Loiseau_Noack_Brunton_2018}. 

\vspace{-0.5cm}
\subsection{Ensemble \gls{sindy} for uncertainty quantification}
\label{esindy}
\vspace{-0.4cm}
Identifying governing equations from clean data is easier since patterns are not masked by poor-quality measurements. In more realistic scenarios, however, \gls{sindy} may fail to identify the correct equations and can instead introduce spurious terms rather than identifying the true system dynamics. These limitations have driven several extensions, including ensemble approaches, E-SINDy~\cite{Fasel2022}, weak formulations, Weak-SINDy~\cite{messenger2021, reinbold2020}, and other extensions to the original algorithm~\cite{Schaeffer2017, alan2021, ali2023} that improve \gls{sindy}'s robustness to noise and limited data.

In this work, we focus on \gls{es} because it quantifies epistemic uncertainty by training $B$ \gls{sindy} models on bootstrap resamples of the data:
\begin{equation}
  \dot{\mathbf{X}}^{(k)} = \Theta^{(k)}\!\left(\mathbf{X}^{(k)}\right)\Xi^{(k)},
  \qquad k = 1, \ldots, B.
\end{equation}
Aggregating across the ensemble yields the mean and epistemic variance (computed element-wise):%
\begin{align}
  \bar{\Xi} &= \frac{1}{B} \sum_{k=1}^{B} \Xi^{(k)}, \\
  \bigl[\sigma^{2}_{\mathrm{ep}}(\Xi)\bigr]_{ij}
  &= \frac{1}{B} \sum_{k=1}^{B}
    \!\left( \Xi^{(k)}_{ij} - \bar{\Xi}_{ij} \right)^{2}.
\end{align}
These statistics propagate to the predicted dynamics $\mathbf{f}(\mathbf{x}) = \Theta(\mathbf{x})\Xi$, giving
\begin{align}
  \sigma^{2}_{\mathrm{ep}}(\mathbf{x})
  &= \frac{1}{B} \sum_{k=1}^{B}
    \left\| \Theta(\mathbf{x})\!\left(\Xi^{(k)} - \bar{\Xi}\right)
    \right\|_{F}^{2},
\end{align}
where $\|\cdot\|_{F}$ denotes the Frobenius norm. Finally, the inclusion probability of the $j$-th library term,
\begin{equation}
  p_j = \frac{1}{B} \sum_{k=1}^{B} \mathbf{1}\!\left( \Xi^{(k)}_j \neq 0 \right),
\end{equation}
enables sparsity promotion: terms with $p_j < \tau$ are discarded, where $\tau \in [0,1]$ is a user-defined threshold, analogous to the thresholding step in STLSQ.
\vspace{-0.5cm}

\subsection{Intelligent sampling for system identification}
\label{inteligentsampling}
\vspace{-0.4cm}
Predictive uncertainty is conventionally separated into two components. \textit{Epistemic uncertainty} reflects model limitations and is reducible in principle, typically estimated through the disagreement among an ensemble of models. \textit{Aleatoric uncertainty} reflects irreducible noise and randomness inherent to the environment, making it substantially harder to quantify. The former is a more reliable guide for finding areas that have not yet been explored and contain meaningful information.

Ensemble methods naturally enable a query-by-committee (QbC) strategy~\cite{seung1992, melville2004diverse}, a concept introduced in 1992 that remains simple and effective. The central idea is to use disagreement among ensemble members as a selection criterion, under the assumption that disagreement correlates positively with model error. In the low-data regime, committee members generate distinct hypotheses, and their disagreement identifies regions of high epistemic uncertainty. Formally, given a candidate set of initial conditions
$\mathcal{X}_{\mathrm{cand}} = \{\mathbf{x}_1, \ldots, \mathbf{x}_M\}$,
the QbC acquisition function selects the next experiment as
\begin{equation}
  \mathbf{x}^* = \underset{\mathbf{x} \in \mathcal{X}_{\mathrm{cand}}}{\mathrm{arg\,max}}
  \; \sigma^{2}_{\mathrm{ep}}(\mathbf{x}),
\end{equation}
where $\sigma^{2}_{\mathrm{ep}}(\mathbf{x})$ quantifies committee disagreement
on the predicted dynamics at $\mathbf{x}$.
A new trajectory is then collected from $\mathbf{x}^*$ and appended to the
training set, the ensemble is retrained, and the process repeats until a
stopping criterion is met.

When learning ODE dynamics from data, epistemic uncertainty propagates along trajectories and is shaped by uncertainty over model parameters and initial conditions. Unlike PDEs, where acquisition must account for spatiotemporal structure, ODEs reduce the problem to selecting along a single trajectory, providing a more tractable setting for developing selective acquisition strategies. This makes QbC particularly well suited for active learning under limited data~\cite{settles2012}, where selecting informative initial conditions reduces the number of experiments needed to identify the underlying dynamics.

Moving from \glspl{ode} to \glspl{pde} substantially increases the difficulty of sampling. Varying initial conditions and training on the resulting trajectories often introduces redundant temporal information rather than new dynamical structure, since in various \gls{pde} systems trajectories from nearby initial conditions cover similar regions of the state space.

Several sampling strategies have been proposed in the literature. When the objective is identifying governing dynamics, the choice of spatiotemporal snapshots becomes critical, as sampling multiple initial conditions may not outperform random selection~\cite{Fasel2022}. Some methods acquire full trajectories by selecting initial conditions from a pool~\cite{musekamp2024}, without targeting the ultra-low-data regime. Others query classical solvers using informative initial conditions and \gls{pde} parameters~\cite{musekamp2024, gao2023}, improving predictive accuracy but not addressing equation discovery.
\vspace{-0.6cm}

\section{Methods}
\label{methods}
\vspace{-0.4cm}
This section explains how to combine active learning with \gls{es} in the very-low-data regime, assuming that collecting new data samples is costly. We also discuss different convergence criteria that are used with active learning.
\vspace{-0.5cm}

\begin{figure}[t!]
    \begin{center}
        \includegraphics[width=\textwidth]{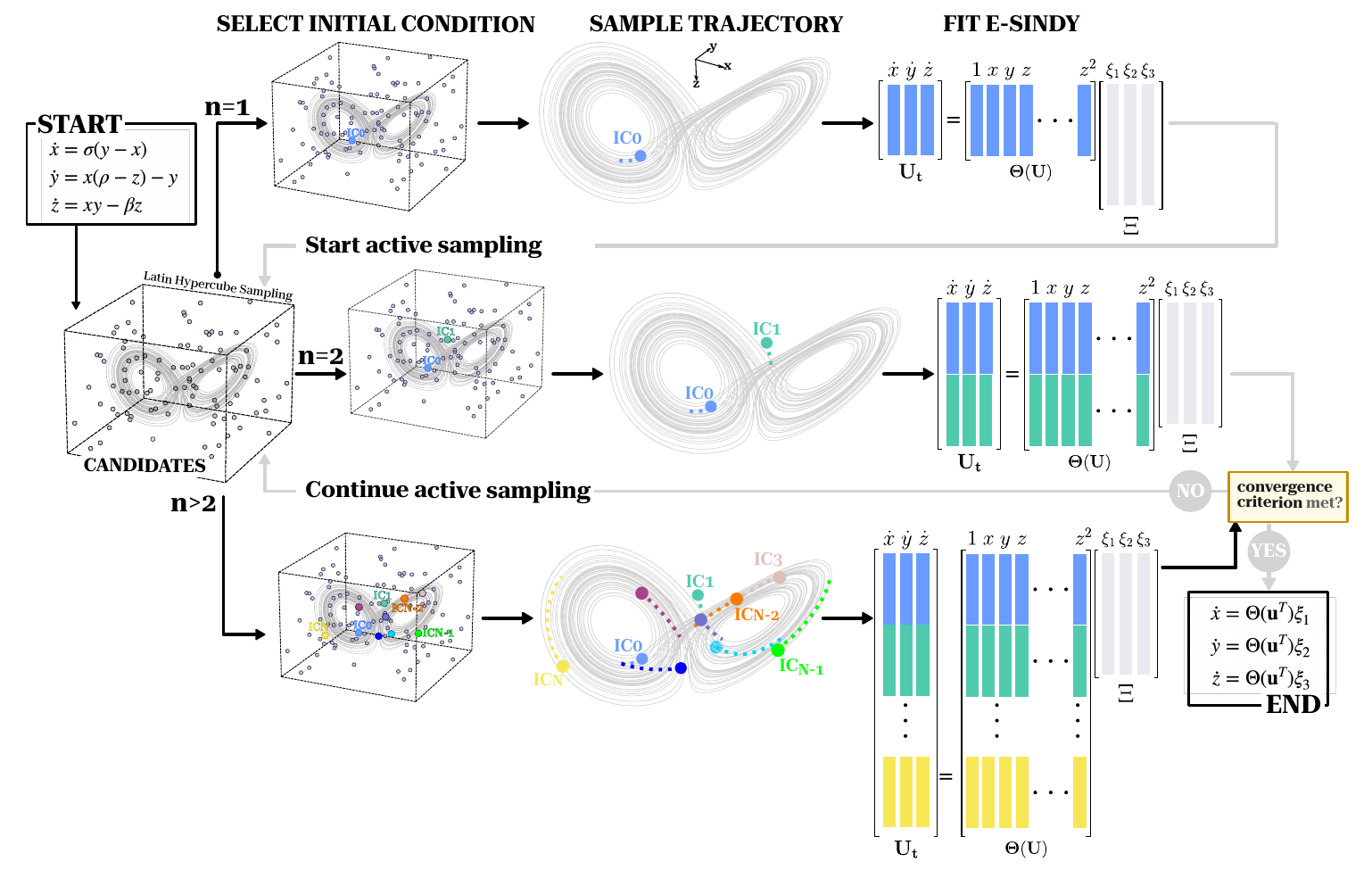}
        \vspace*{-0.5cm}
        \caption{Active learning loop with \gls{es} for \gls{ode} dynamics discovery. Using the Lorenz system as an example, candidate initial conditions are sampled around the attractor and used to initialize short trajectories. An ensemble of SINDy models is trained on the current dataset, and the candidate with the highest predictive uncertainty is iteratively added to the training set until convergence, defined as agreement among all ensemble members on both the predictive error and values of the active coefficients.}
        \label{fig1:main_pipeflow}
        \vspace{-0.7cm}
    \end{center}
\end{figure}

\subsection{Towards the ultra-low-data limit}
\label{ultralowdata}
\vspace{-0.4cm}
Capturing complex, time-dependent behavior with minimal measurements requires balancing the amount of sampled information with the ability to represent the system states~\cite{themis2018, champion2019}. Our objective is to develop a strategy for determining \textit{where} and \textit{when} to sample so that each new data point maximally guides dynamics discovery. For \gls{ode}, this corresponds to selecting informative initial conditions from which short trajectories are generated; for \gls{pde}, it involves identifying spatiotemporal locations that provide the greatest insight into the governing dynamics.

\begin{figure}[t!]
    \begin{center}
        \includegraphics[width=1\textwidth]{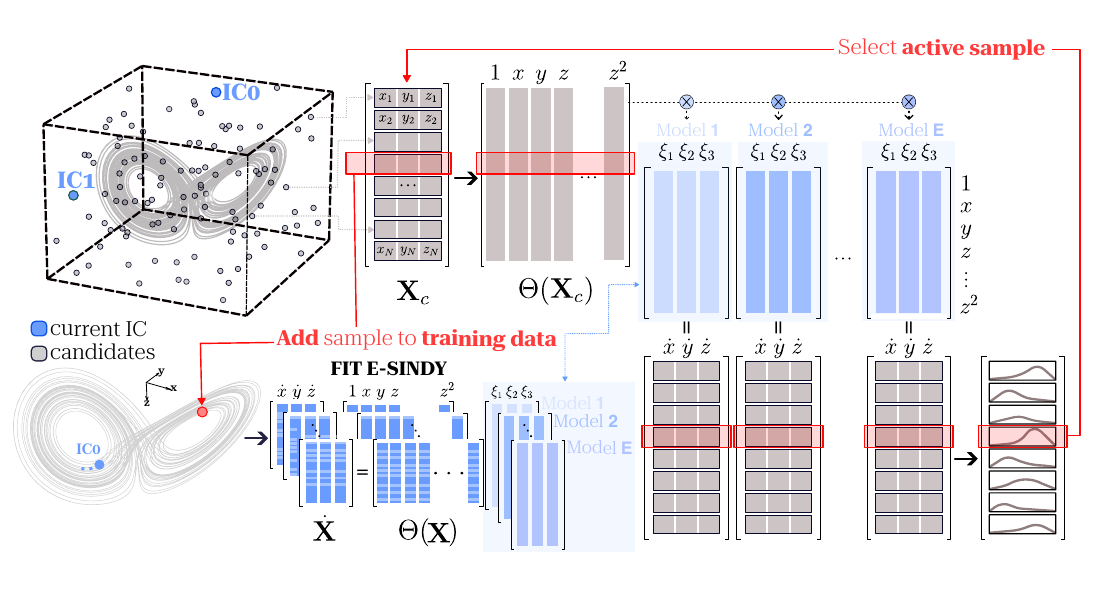}
        \vspace*{-10mm}
        \caption{Detailed description of the active learning sampling strategy outlined in the intermediate steps of Fig.~\ref{fig1:main_pipeflow}. Initial conditions are selected iteratively from a pool of candidates based on ensemble disagreement, prioritizing those where the epistemic uncertainty of the E-SINDy model is highest, as these carry the most information about the governing dynamics. Once a candidate is selected, a short trajectory segment is integrated from that initial condition and appended to the training dataset, which is used to refit the E-SINDy model. This procedure is repeated until convergence, defined here as agreement among all ensemble members in their predictions, ensuring a robust and accurate sparse identification. The approach is computationally efficient relative to fully Bayesian alternatives: E-SINDy training is fast and candidate evaluation reduces to a matrix multiplication, making the active learning loop tractable even for large candidate pools.}
        \label{Fig:active_learning}
        \vspace*{-0.8cm}
      \end{center}
\end{figure}

More precisely, the goal of this work is to learn a dynamical system
\begin{align}
    \frac{\mathrm{d}\mathbf{x}}{\mathrm{d}t} = \mathbf{f}(\mathbf{x})
\end{align}
from as few data samples $\mathbf{x}_1, \mathbf{x}_2, \ldots, \mathbf{x}_m$ as possible. To this end, we \emph{actively} select these samples to maximally inform model discovery, as illustrated schematically in Fig.~\ref{fig1:main_pipeflow}. We build on the sparse identification of nonlinear dynamics (SINDy) and ensemble-SINDy frameworks, which approximate $\mathbf{f}(\mathbf{x})$ as a sparse expansion in a library of candidate linear and nonlinear functions:
\begin{align}
    \frac{\mathrm{d}\mathbf{x}}{\mathrm{d}t}  = \mathbf{f}(\mathbf{x})\approx \begin{bmatrix} \theta_1(\mathbf{x}) & \theta_2(\mathbf{x}) & \cdots& \theta_p(\mathbf{x}) \end{bmatrix} \boldsymbol{\Xi}
\end{align}
where the coefficient matrix $\boldsymbol{\Xi}$ is made as sparse as possible while still capturing the observed dynamics.

\vspace{-0.3cm}
\subsubsection{Active-SINDy: Sampling for \gls{ode}}
\vspace{-0.4cm}
Following Fig.~\ref{fig1:main_pipeflow}, the proposed active learning strategy selects initial conditions that maximize discrimination among competing models in \gls{es}. Regions of high predictive disagreement indicate poorly resolved dynamics and therefore represent the most informative directions for additional data acquisition. 

The epistemic uncertainty, used as acquisition function, is quantified as the variance of the ensemble predictions obtained from $B$ bootstrapped SINDy models $\boldsymbol{\Xi}^{(k)}$ (see Eq. 7). To maintain diversity, candidate initial conditions that lie close to existing samples, or that are duplicates, are removed. Eliminated candidates are replaced to preserve a fixed pool size
N, treated as a hyperparameter. Because ensemble evaluation occurs after training, computing predictive distribution over candidate points is considerably less expensive than retraining the models with additional data. The selection criterion therefore focuses on model disagreement rather than predictive accuracy. Details of the full logic are described in algorithm~\ref{alg:active_sindy}.

\begin{algorithm}[b!]
    \caption{Active initial condition selection combined with E-SINDy for ODEs}
    \label{alg:active_sindy}
    \begin{algorithmic}[1]
    \Require Candidate pool $\mathcal{C} = \{\mathbf{x}_i\}_{i=1}^{N}$, training set $\mathcal{D}$, ensemble $\{\Xi^{(k)}\}_{k=1}^{B}$, radius $\rho$
    \Ensure Updated $\mathcal{D}$, retrained ensemble
    \State \textbf{Rank.} Score each $\mathbf{x}_i \in \mathcal{C}$ by $s(\mathbf{x}_i) = \frac{1}{B}\sum_{k=1}^{B}\left\|\Theta(\mathbf{x}_i)\bigl(\Xi^{(k)} - \hat{\boldsymbol{\xi}}\bigr)\right\|^2$, $\;\hat{\boldsymbol{\xi}} = \operatorname{median}_k\{\Xi^{(k)}\}$; sort $\mathcal{C}$ descending.
    \State \textbf{Filter.} Remove $\mathbf{x}_i$ if $\exists\,\mathbf{x}_j^{*} \in \mathcal{D}: \|\mathbf{x}_i - \mathbf{x}_j^{*}\|_2 < \rho$; refill to $|\mathcal{C}|=N$.
    \State \textbf{Update.} Set $\mathbf{x}^{*} = \operatorname{arg\,max}_{\mathbf{x}_i \in \mathcal{C}_{\mathrm{filt}}} s(\mathbf{x}_i)$, update $\mathcal{D} \leftarrow \mathcal{D} \cup \{\mathbf{x}^{*}\}$, retrain $\{\Xi^{(k)}\}_{k=1}^{B}$.
    \State Repeat Steps~1--3 until max iterations or $\hat{\boldsymbol{\xi}}$ converges.
    \end{algorithmic}
\end{algorithm}

Although additional criteria, such as leverage scores~\cite{Rudi2018}, can be incorporated to assess the information contained in the library matrix in the \gls{ode} setting, but are ill-suited for ultra-low-data application where no representative subset is yet available to analyze. Methods based on eigenvalue analysis~\cite{Pukelsheim2006} or QR factorization~\cite{Manohar2018} typically rely on large candidate pools, whereas the present framework is designed to begin with minimal data and expand iteratively in a computationally efficient manner; see Fig.~\ref{Fig:active_learning}.

In this work, the sampling procedure is initialized using a Latin Hypercube Sampling (LHS) to generate a space-filling set of candidate initial conditions around the attractor. LHS defines a well-distributed sampling domain from which subsets of candidates are drawn for active learning evaluation. This ensures broad coverage of the relevant state space while allowing the ensemble-based uncertainty criterion to identify the most informative points within a diverse candidate pool (see Fig.~\ref{Fig:active_learning}).

\begin{figure}[t!]
    \begin{center}
        \includegraphics[width=0.6\textwidth]{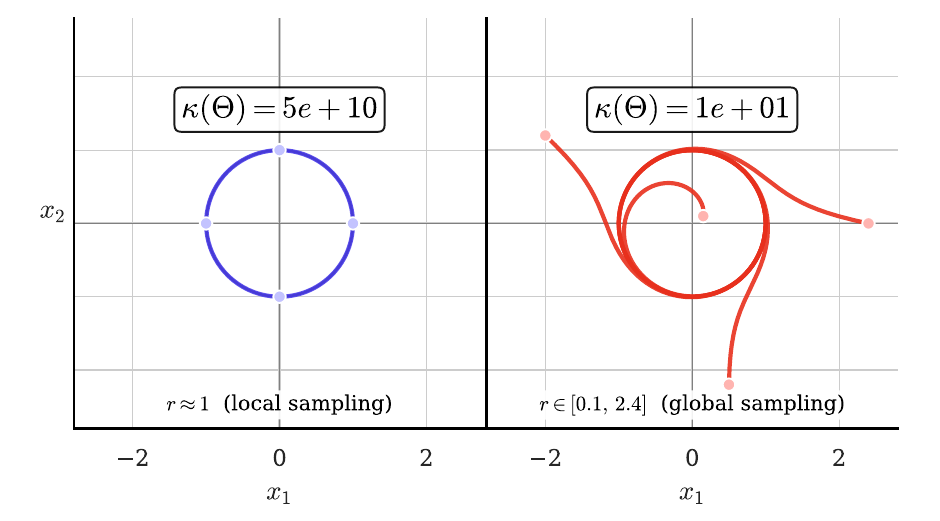}
    \end{center}
    \vspace*{-8mm}
    \caption{Sampling strategy and its effect on the condition number $\kappa(\Theta)$ of the library matrix for the Hopf normal form oscillator. Sampling restricted to the limit cycle ($r \approx 1$, blue) yields a poorly conditioned library matrix ($\kappa(\Theta) = 5\times10^{10}$), as the trajectories explore only a narrow region of the state space, failing to capture the nonlinear dynamics away from the attractor. Sampling across a broader range of initial conditions ($r \in [0.1, 2.4]$, red) reduces the condition number by nine orders of magnitude ($\kappa(\Theta) = 1\times10^{1}$), enabling the algorithm to resolve the transient dynamics and disambiguate the governing nonlinearities. This illustrates that even seemingly simple systems present nontrivial sampling challenges.}
    \label{hopf_sampling_fair}
    \vspace{-0.5cm}
\end{figure}

If the data is collected close to the attractor, the global vector field will be poorly constrained and nonlinear terms will be unidentifiable under measurement noise. For instance, consider the normal-form Hopf oscillator
\begin{equation}
    \dot{x}_1 = \mu x_1 - x_2 - x_1(x_1^2+x_2^2), \quad
    \dot{x}_2 = x_1 + \mu x_2 - x_2(x_1^2+x_2^2),
    \label{eq:hopf}
\end{equation}
with stable limit cycle $r=\sqrt{\mu}$. On the attractor, the library condition number reaches $\sim\!10^{10}$, making coefficient recovery ill-posed. Initial conditions distributed across phase space reduce the condition number to $\mathcal{O}(10)$ and recover the true nonlinear structure (see Fig.~\ref{hopf_sampling_fair}). The acquisition function $\sigma^2_\mathrm{ep}(\mathbf{x})$ exploits ensemble disagreement to focus sampling toward these globally informative regions.

\vspace{-0.3cm}
\subsubsection{Active-SINDy: Extension to \gls{pde}}
\begin{figure}[t]
    \begin{center}
        \includegraphics[width=0.9\textwidth]{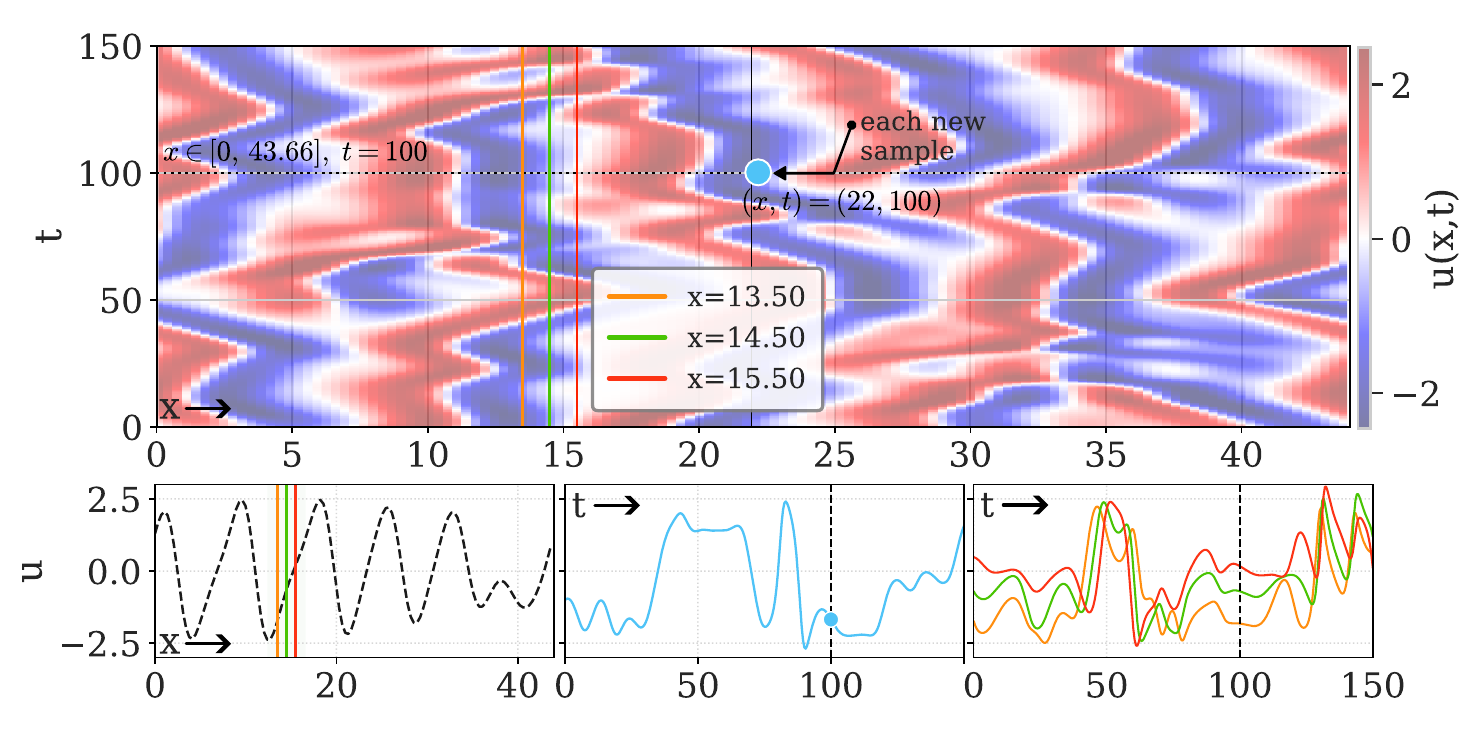}
    \end{center}
    \vspace{-0.5cm}
    \caption{Illustrative sampling strategies for the Kuramoto-Sivashinsky equation. Conventional approaches, including prior E-SINDy implementations, utilize the entire spatiotemporal domain, incorporating thousands of redundant points. Fixing a spatial location and sampling the full time series (orange, green, and red vertical lines) reduces the dataset size but retains substantial redundancy, as nearby spatial locations exhibit highly correlated dynamics (bottom right panel). In contrast, a single spatiotemporal point (cyan dot, $(x,t) = (22, 100)$) can carry sufficient information to identify the governing dynamics, motivating the ultra-low data sampling strategy developed in the present work. }
    \label{pdeexplanation}
    \vspace{-0.5cm}
\end{figure}

\vspace{-0.4cm}
The PDE setting gives direct access to the full spatiotemporal field, so rather than selecting initial conditions as in the ODE case, the method queries individual locations  (see Fig.~\ref{pdeexplanation}). At each iteration a single point $(x_i, t_i)$ is selected from the $M = N_x \times N_t$ grid to maximize information for identifying $\boldsymbol{\xi}$ in $\partial_t u = \Theta(u)\,\boldsymbol{\xi}$. Spatial derivatives in $\Theta(u) \in \mathbb{R}^{M \times p}$ are computed via the FFT and $\partial_t u$ by spectral differentiation in time. Library columns are normalized to unit variance,
\begin{equation}
    \tilde{\theta}_{ij} = \theta_{ij}/\sigma_j, \qquad \sigma_j = \mathrm{std}_{i \in \mathcal{T}}[\theta_{ij}] + \varepsilon,
    \label{eq:col_normalise}
\end{equation}
to reduce collinearity before sparse regression.

A bootstrap ensemble of $B$ \gls{sindy} models is maintained throughout the loop. Each member is fit by \gls{stlsq} on a random subsample of $\max(p+2,\lfloor 0.8|\mathcal{S}|\rfloor)$ points drawn with replacement from the current training set $\mathcal{S}$, with an $L_2$ penalty scaled by the condition number of $\tilde{\Theta}_\mathcal{S}$ to stabilize the fit when the system is underdetermined. A small \gls{stlsq} threshold $\tau_\mathrm{ens}$ is used, so that different bootstrap samples can select different active terms. The ensemble mean $\bar{\boldsymbol{\xi}}$ is evaluated on a held-out test set at each iteration. Let $g = p^{-1}\sum_{j}|\bar{\xi}_j|$ denote the mean magnitude of the ensemble-mean coefficients, and let $\mathcal{A} = \{j : |\bar{\xi}_j| > 0.05\, g\}$ be the set of active terms. Convergence is monitored through the mean coefficient of variation over $\mathcal{A}$,
\begin{equation}
    \rho_\mathrm{cov} = \frac{1}{|\mathcal{A}|}\sum_{j\in\mathcal{A}} \frac{\mathrm{std}_k[\hat{\xi}_j^{(k)}]}{\max(|\bar{\xi}_j|,\; 0.01\, g)},
    \label{eq:coef_cov}
\end{equation}
where $0.01\, g$ acts as a lower bound on the denominator to keep small coefficients from inflating the ratio. The loop terminates when $\rho_\mathrm{cov} < 5\%$ or the sampling budget $N$ is reached.

A new point is chosen by one of two acquisition functions. The D-optimal score selects the point that maximizes the marginal log-determinant gain of $\tilde{\Theta}_\mathcal{S}^\top\tilde{\Theta}_\mathcal{S}$; by the matrix determinant lemma,
\begin{equation}
    s_i^{\mathrm{dopt}} = \log(1 + h_i), \qquad h_i = \tilde{\boldsymbol{\theta}}_i^\top\!\left(\tilde{\Theta}_\mathcal{S}^\top\tilde{\Theta}_\mathcal{S} + \lambda\mathbf{I}\right)^{-1}\!\tilde{\boldsymbol{\theta}}_i,
    \label{eq:dopt_score}
\end{equation}
with $\lambda$ a small ridge term proportional to the mean diagonal of $\tilde{\Theta}_\mathcal{S}^\top\tilde{\Theta}_\mathcal{S}$. The ensemble score combines model error and ensemble disagreement,
\begin{equation}
    s_i^\mathrm{ens} = \tilde{\mu}_i \cdot \tilde{\rho}_i, \qquad \mu_i = \frac{1}{B}\sum_k r_i^{(k)}, \quad \rho_i = \frac{\mathrm{std}_k[r_i^{(k)}]}{\mu_i + \varepsilon},
    \label{eq:ens_score}
\end{equation}
where $r_i^{(k)} = |\partial_t u_i - \tilde{\boldsymbol{\theta}}_i^\top\hat{\boldsymbol{\xi}}^{(k)}|$ and $\tilde{\cdot}$ denotes min-max normalization to $[0,1]$. High scores mark points where the model fits poorly and the ensemble disagrees.

To prevent clustering, a candidate $(x_i,t_i)$ is retained only if its normalized grid distance to every point in $\mathcal{S}$ exceeds $\delta_{\min}$. The highest-scoring surviving candidate is added; if none survive, the global $\operatorname{arg\,max}_i s_i$ is added unconditionally. The initial set $\mathcal{S}_0$ is seeded by LHS under the same filter. Once the loop terminates, a final \gls{stlsq} fit is performed on $\mathcal{S}$ with the sparsity threshold $\tau^\star$ chosen  to minimize the Bayesian Information Criterion (BIC), which trades off fit quality against model complexity,
\begin{equation}
    \mathrm{BIC}(\tau) = n_\mathrm{train}\log\hat{\sigma}^2(\tau) + k(\tau)\log n_\mathrm{train}.
    \label{eq:bic}
\end{equation}
where $\hat{\sigma}^2$ is the test-set mean squared residual and $k(\tau)$ the number of active coefficients. The fitted coefficients are rescaled by $\boldsymbol{\sigma}$ to recover the physical coefficients $\hat{\boldsymbol{\xi}}$. The full procedure is given in Algorithm~\ref{alg:active_sindy_pde}.

\begin{algorithm}[t]
    \caption{Active space-time point selection combined with \gls{es} for \glspl{pde}}
    \label{alg:active_sindy_pde}
    \begin{algorithmic}[1]
    \Require Spatiotemporal grid $\{(x_i,\,t_i,\,\partial_t u_i,\,\tilde{\boldsymbol{\theta}}_i)\}_{i=1}^{M}$, budget $N$, acquisition mode \texttt{acq}, ensemble $\{\hat{\boldsymbol{\xi}}^{(k)}\}_{k=1}^{B}$, threshold $\tau_{\mathrm{ens}}$, diversity threshold $\delta_{\min}$
    \Ensure Sparse coefficient vector $\hat{\boldsymbol{\xi}}$
    \State \textbf{Initialize.} $\mathcal{S} \leftarrow \mathrm{LHS}(N_0)$, $N_0 = \min(p+2,\,20)$, filtered by diversity criterion
    \State \textbf{Score.} Score each $(x_i,t_i) \notin \mathcal{S}$ by $s_i = s_i^{\mathrm{dopt}}$ or $s_i = s_i^{\mathrm{ens}}$ according to \texttt{acq}; sort descending
    \State \textbf{Filter.} Remove $(x_i,t_i)$ if $\exists\,(x_j,t_j) \in \mathcal{S}: d(i,j) < \delta_{\min}$; fall back to global $\operatorname{arg\,max}_i\,s_i$ if all are removed
    \State \textbf{Update.} Set $(x^\star,t^\star) = \operatorname{arg\,max}_{i \in \mathcal{C}_{\mathrm{filt}}}\,s_i$, update $\mathcal{S} \leftarrow \mathcal{S} \cup \{(x^\star,t^\star)\}$, retrain $\{\hat{\boldsymbol{\xi}}^{(k)}\}_{k=1}^{B}$ via \gls{stlsq} with threshold $\tau_{\mathrm{ens}}$
    \State Repeat Steps~2--4 until $|\mathcal{S}| = N$
    \State \textbf{Identify.} Run \gls{stlsq} on $\mathcal{S}$; select $\tau^\star$ by minimizing \eqref{eq:bic}; return $\hat{\boldsymbol{\xi}} = \hat{\boldsymbol{\xi}}^{\star}_{\mathrm{sc}} \oslash \boldsymbol{\sigma}$
    \end{algorithmic}
\end{algorithm}

\vspace{-0.4cm}
\section{Results \& Discussion}
\label{Results & Discussion}
\vspace{-0.4cm}
Here we present the results of the active learning strategy on the Lorenz system, a canonical benchmark for dynamical system identification, and on two partial differential equations (\gls{pde}) with different complexity. Performance is compared against random sampling in all cases, and against a D-optimality criterion for the \gls{pde} examples, where sampling in the ultra-low-data regime is particularly challenging.
\vspace{-0.4cm}
\subsection{Active-SINDy: Sampling for \gls{ode}}
\label{resODE}
\vspace{-0.4cm}

The Lorenz system was originally introduced as a reduced model of atmospheric convection and is described by the nonlinear system
\begin{equation}
    \begin{aligned}
    \dot{x} &= \sigma (y - x), \
    \dot{y} &= x(\rho - z) - y, \
    \dot{z} &= xy - \beta z,
    \end{aligned}
    \label{eq:lorenz}
\end{equation}
where $\sigma$, $\rho$, and $\beta$ are positive parameters. For the classical parameter values, which are used in this work, $\sigma = 10$, $\rho = 28$, and $\beta = 8/3$, the system exhibits chaotic dynamics. Due to its well-understood structure and chaotic behavior, the Lorenz system is widely used as a benchmark for dynamical system identification methods

\begin{figure}[t!]
    \vspace{-0.9cm}
    \begin{center}
        \includegraphics[width=1\textwidth]{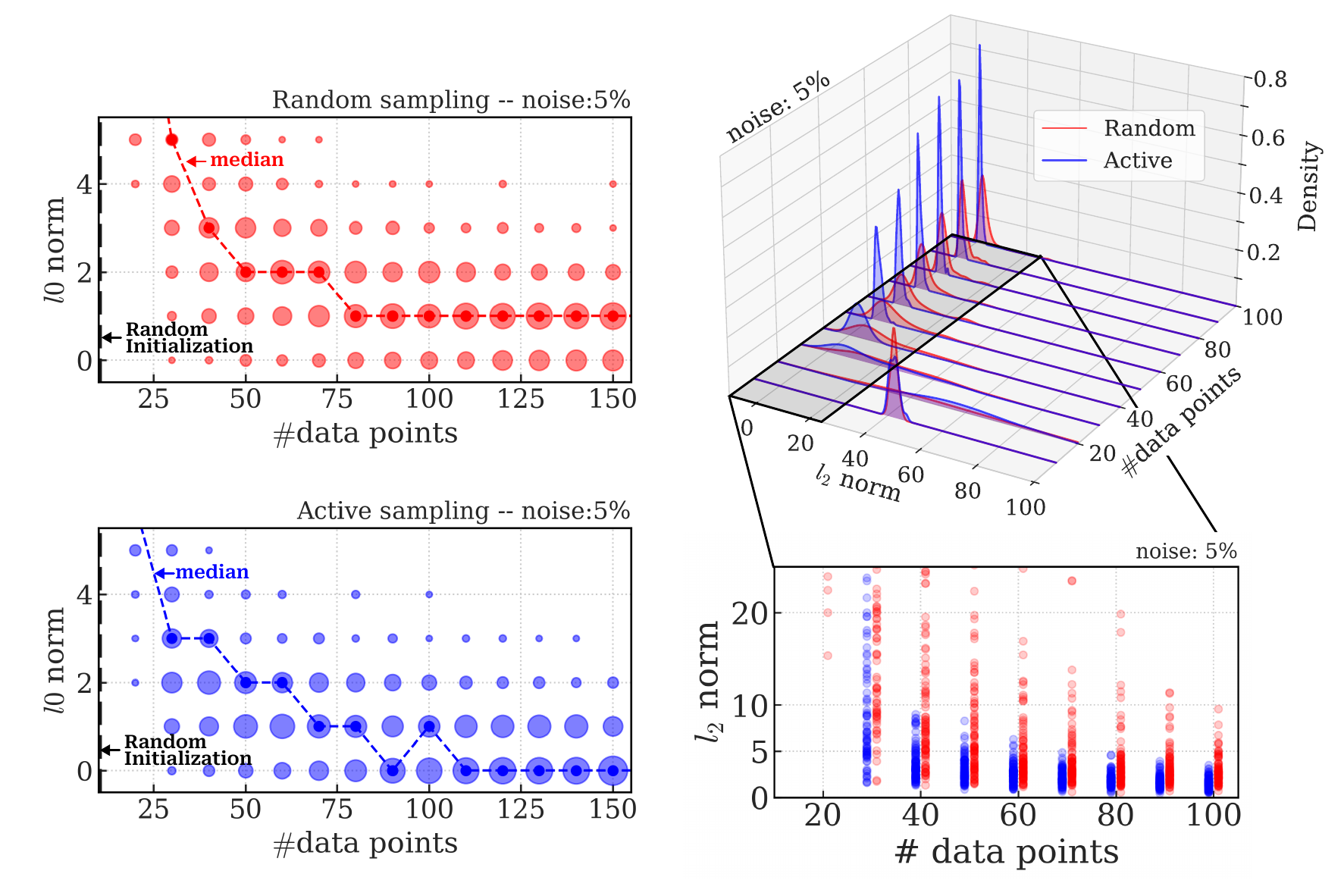}
    \end{center}
    \vspace{-0.8cm}
    \caption{Comparison between active learning (blue) and random sampling (red) for dynamics discovery of the chaotic Lorenz system under 5\% noise, evaluated via the $\ell_0$ and $\ell_2$ norms in the ultra-low data limit. The left panels show the median performance (dashed line) over 100 independent repetitions of the full loop, accounting for the variability introduced by the random selection of the initial candidate pool. The upper right panel shows a 3D representation of the frequency distribution of the minimum number of points required to achieve correct sparse identification across repetitions, as measured by the $\ell_0$ norm. The lower right panel shows the $\ell_2$ norm as a function of the data budget, illustrating the improved robustness and data efficiency of active learning over random sampling.}
    \vspace{-0.3cm}
    \label{fig:ode-results-1}
\end{figure}

Following the active learning strategy introduced in Fig.~\ref{Fig:active_learning}, Fig.~\ref{fig:ode-results-1} shows the $\ell_0$ and $\ell_2$ norms obtained for the Lorenz system with additive Gaussian noise, where the noise standard deviation was set to $5\%$ of the root-mean-square value of the training data. Compared with random sampling, the active learning approach recovers the correct governing equations using only 100 data points distributed across 10 trajectories. The correct coefficients are also identified earlier in the sampling procedure, as indicated by the tighter clustering of the blue points in the $\ell_2$ norm, which measures the total coefficient error over the library terms. The three-dimensional plots further show that, even with only 40 data points, the active learning procedure already captures the correct dynamics despite the presence of noise in the training data. 

\begin{figure}[t!]
    \begin{center}
        \includegraphics[width=1\textwidth]{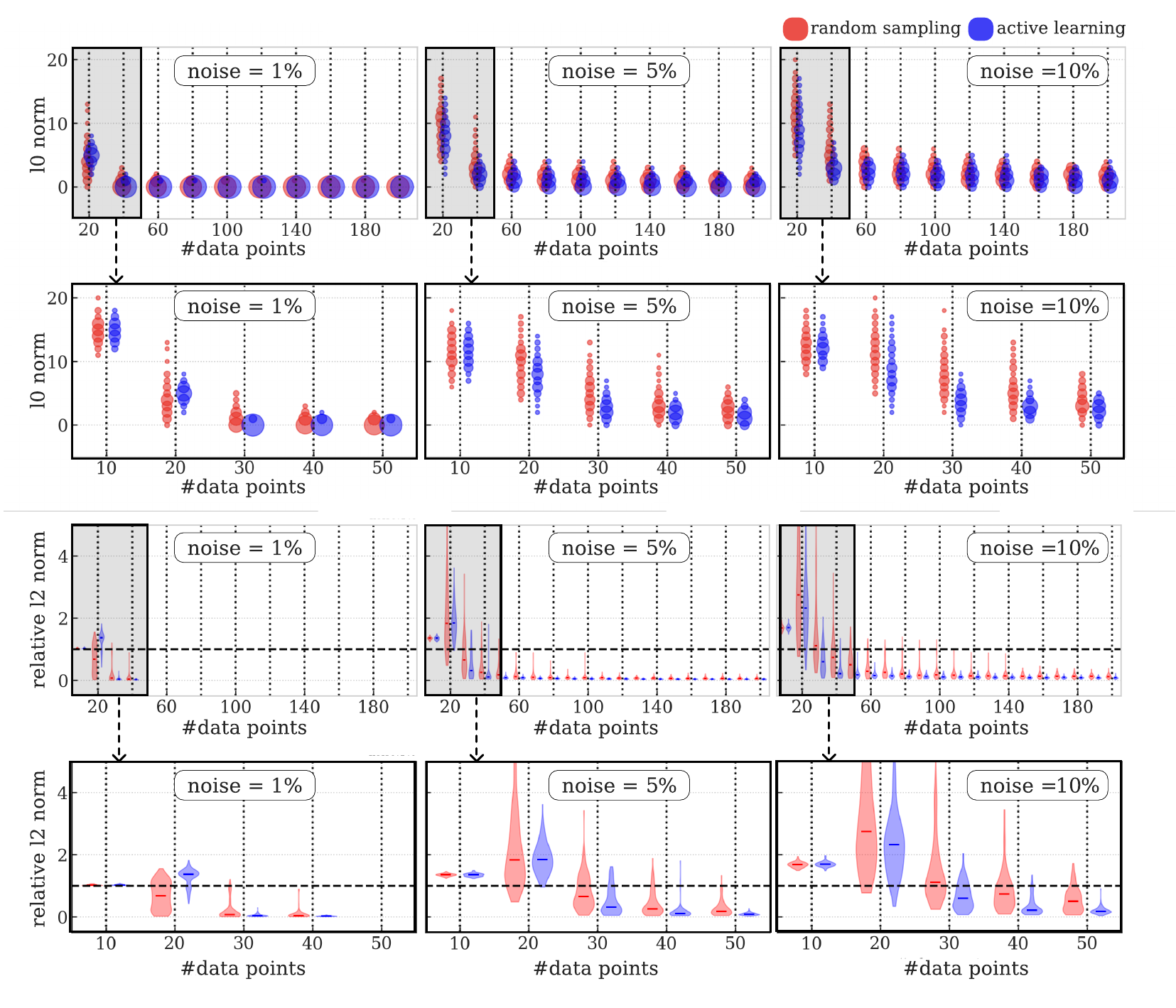}
    \end{center}
    \vspace{-1cm}
    \caption{Evaluation of the noise effect on the active learning sampling strategy (blue) compared to random sampling (red) for the Lorenz system, across 1\% (left), 5\% (middle), and 10\% (right) additive Gaussian noise levels. The first row shows the $\ell_0$ norm over the longer data budget range to show convergence, while the second row focuses on the ultra-low data limit (0 to 50 points). The third and fourth rows show the relative $\ell_2$ norm, defined as the $\ell_2$ norm of the identified coefficients normalized by the norm of the true coefficients evaluated over the entire library matrix, again with the fourth row restricted to the ultra-low data regime.}
    \vspace{-0.5cm}
    \label{fig:ode-results-2}
\end{figure}

\textbf{Effect of noise}. To evaluate the robustness of the method under different noise levels, the same error metrics were evaluated for noise levels of $1\%$, $5\%$, and $10\%$ (see Fig.~\ref{fig:ode-results-2}). Although higher noise levels may be addressed through weak-form \cite{messenger2021} ensemble approaches, excessive smoothing can remove relevant information in the search space, requiring the support of the test functions to be chosen consistently with the structure of the dataset. In the present low-data setting, where only short trajectories are available, the application of weak-form methods is not straightforward and may require longer time series to fully benefit from the formulation. For this reason, this work focuses on the selection of informative initial conditions from short trajectories. As shown in Fig.~\ref{fig:ode-results-2}, the active learning procedure outperforms random sampling across all noise levels and maintains comparable performance between the $5\%$ and $10\%$ noise cases, finding the correct number of active terms with less than 100 samples with the correct coefficients as well.

Note that the first x-axis location in Fig.~\ref{fig:ode-results-2}, corresponding to 10 data points, does not involve any active learning: this initial trajectory is generated from a randomly chosen initial condition, which is kept the same for both the random sampling and active learning evaluations to ensure a fair comparison. Ten points is selected to ensure that the numerical differentiation remains as accurate as possible, even in the highest-noise cases. Although the $\ell_2$ error already appears small at this point, the $\ell_0$ norm plot reveals many spurious terms, indicating that the correct structure has not yet been identified.

\begin{figure}[t!]
    \setlength{\abovecaptionskip}{0pt}
    \begin{center}
        \includegraphics[width=1\textwidth]{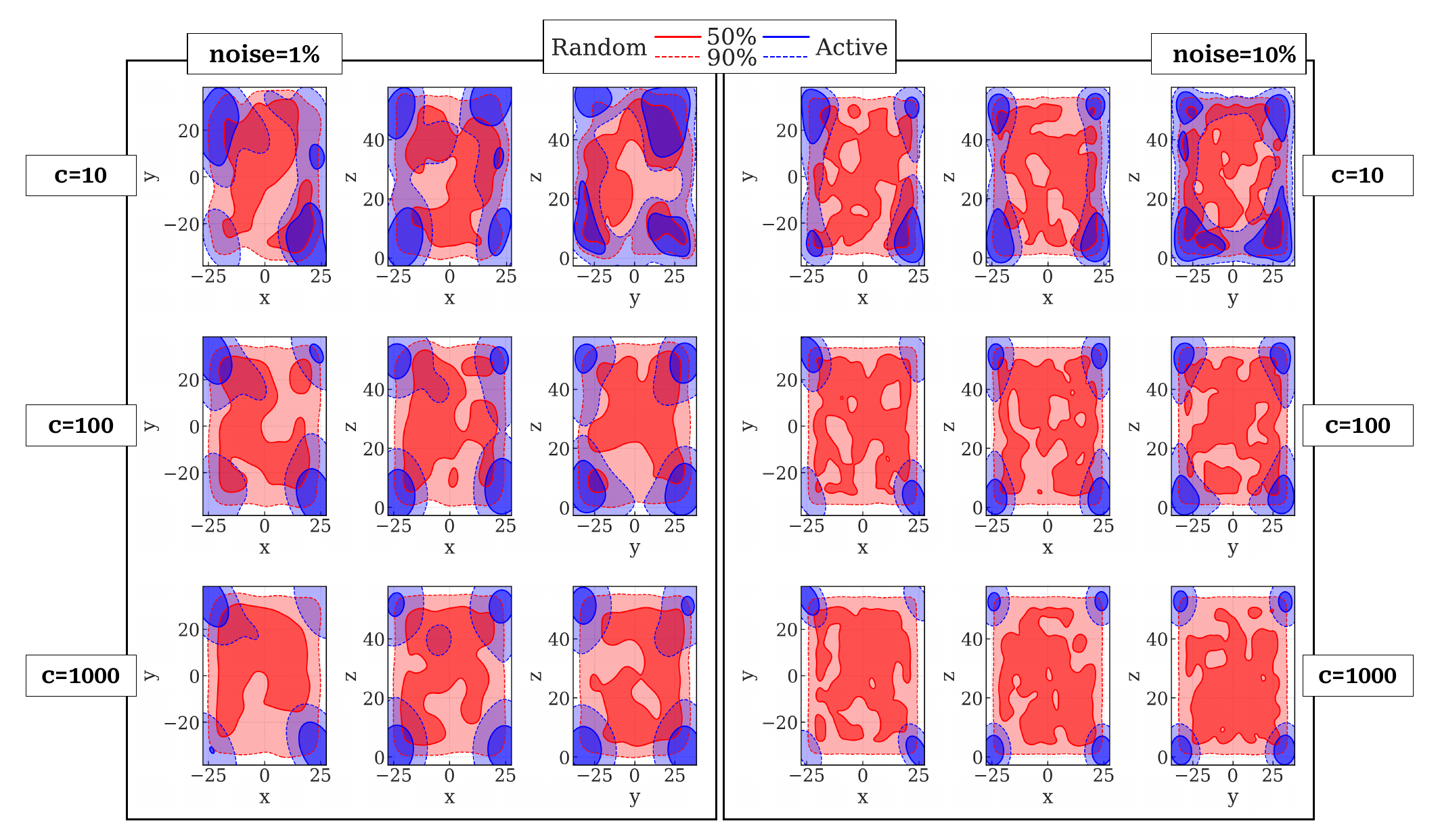}
    \end{center}
    \caption{Location of initial conditions (ICs) in the training dataset for active learning (blue) versus random sampling (red) across different candidate pool sizes ($c$) for the Lorenz system. Solid and dashed contours enclose 50\% and 90\% of the ICs, respectively. Active learning consistently concentrates ICs in localized regions of the state space, in contrast to random sampling, which distributes them uniformly across the domain. Results are shown for two noise levels, 1\% (left) and 10\% (right), and three candidate pool sizes ($c = 10$, top; $c = 100$, middle; $c = 1000$, bottom), demonstrating that the spatial preference of active learning is robust to both noise level and candidate pool size.}
    \label{fig:candidates}
    \vspace{-0.4cm}
\end{figure}

\textbf{Location of sampled initial conditions.} An additional interesting analysis is the spatial distribution of the selected samples relative to the attractor. As previously described, sampling begins by defining a bounding box of prescribed size around the Lorenz attractor, from which a pool of candidate initial conditions is randomly drawn for evaluation. Worth highlighting, independently of the noise level in the dataset, the selected initial conditions are predominantly located near the boundaries of this domain (see Fig.~\ref{fig:candidates})

This observation suggests that the most informative data are not located directly on the attractor, but rather in regions away from it. Traditional sampling strategies for the Lorenz system typically emphasize long trajectories located on the attractor. In contrast, the active learning framework consistently favors trajectories initiated farther from the invariant set, where transient dynamics provide additional structural information about the governing equations.

A similar behavior was observed when applying the method to the nonlinear Hopf oscillator. When sampling is concentrated near the oscillator’s attractor, SINDy identifies only the linearized dynamics and fails to recover the nonlinear terms. In contrast, trajectories initiated sufficiently far from the attractor capture the nonlinear contributions necessary for correct model identification. This highlights a broader principle: for systems admitting simpler local representations near invariant sets, restricting sampling to the attractor may hide the nonlinear structure of the governing equations. Sampling beyond the attractor is therefore essential for disambiguating competing models and recovering the full dynamics.
\vspace{-0.4cm}

\subsection{Active-SINDy: Extension to \gls{pde}}
\label{resPDE}
\vspace{-0.4cm}
When developing an active learning technique to sample a PDE, focusing on maximizing the information gathered from the spatiotemporal domain is especially important and effective in the ultra-low data limit. The advantage over random sampling becomes clear when the data contains a mix of informative and uninformative regions. Fluid flows are a natural example: many cases of practical interest contain localized features such as shocks, vortices, or boundary layers that carry far more information than the surrounding regions, while transient or developing zones contribute comparatively little. Active learning exploits this imbalance by concentrating samples where they matter most. In fully turbulent flows, however, most regions are sufficiently informative to characterize the underlying dynamics, and the gains over random sampling tend to be minor. In this section, we analyze the performance of an active learning technique based on uncertainty estimated by an ensemble of \gls{sindy} models, compared to random sampling and a widely used design-of-experiments strategy known as the D-optimal criterion, using the Burgers and Kuramoto-Sivashinsky (KS) equations as benchmark problems.

\textbf{Burgers}. This equation is a nonlinear \gls{pde} that combines convective and diffusive effects~\cite{burgers1948}, serving as a canonical benchmark for shock formation and propagation:

\begin{equation}
    u_t + uu_x = \nu u_{xx}, \quad x \in [0, L], \quad t > 0, 
    \label{eq:burgers}
\end{equation}

\noindent where the nonlinear term $uu_x$ drives wave steepening and shock formation, and the diffusive term $\nu u_{xx}$ regularizes the solution, with $\nu$ controlling the shock width. For small $\nu$, the solution develops sharp shock fronts, making Burgers a widely used benchmark for problems involving discontinuities and steep gradients. 

\begin{figure}[t]
    \begin{center}
        \vspace*{-2.7cm}
        \includegraphics[width=1\textwidth]{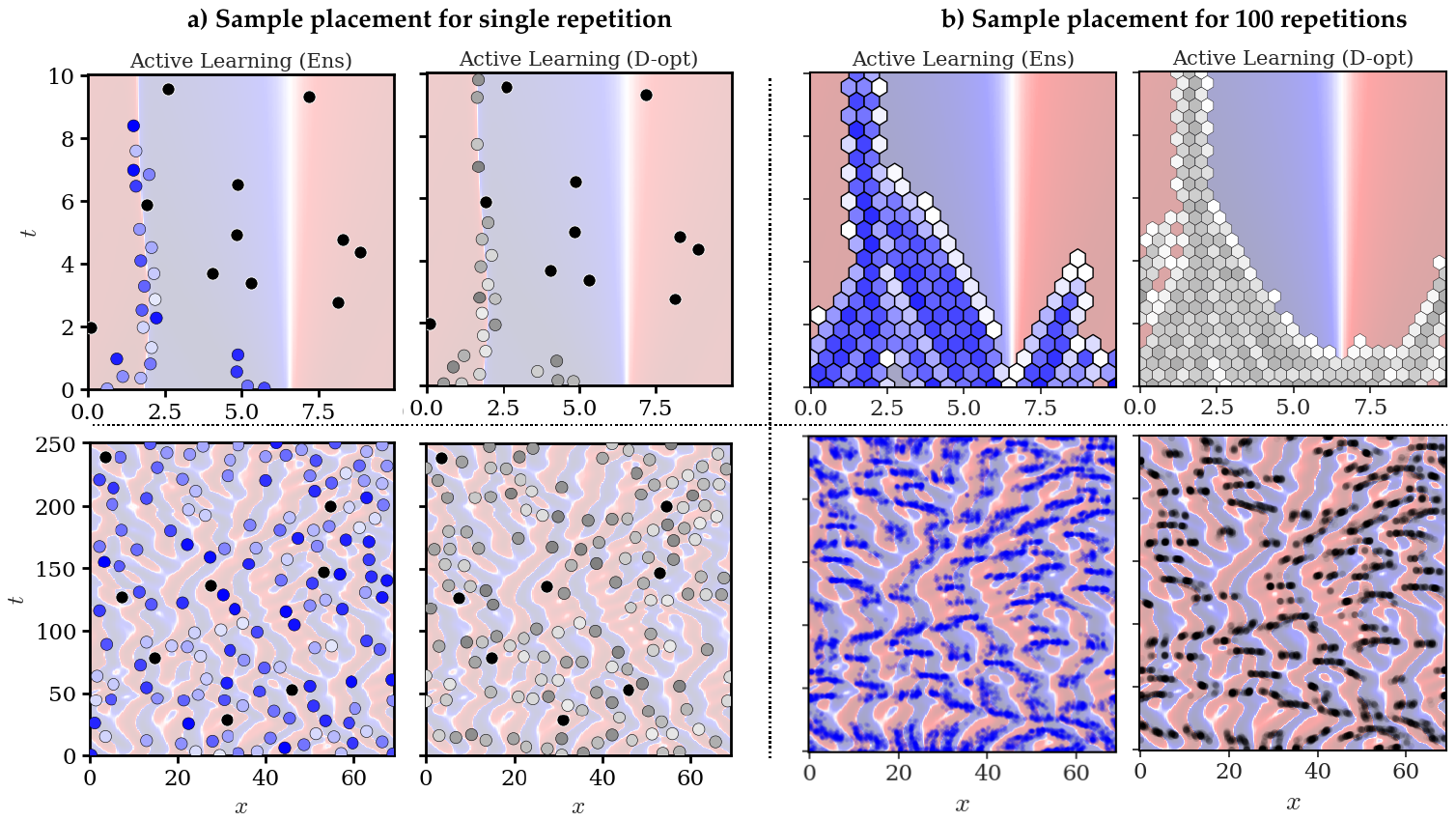}
        \vspace*{-2.9cm}
        \caption{Analysis of sample placement for the two sampling strategies compared in this section: active learning via ensembling (blue) and D-optimal (grey). The left panels show a single representative run with the nominal data budget for Burgers (24 points, top) and KS (100 points, bottom). The right panels show the aggregated sample density over 100 independent repetitions, confirming the consistency of sample placement across runs. In all panels, color intensity reflects sampling frequency, with darker regions indicating higher point density.}
        \vspace*{-0.5cm}
        \label{fig:pderesults-1}
    \end{center}
\end{figure}

\begin{figure}[t]
    \begin{center}
        \vspace*{-2.5cm}
        \includegraphics[width=0.98\textwidth]{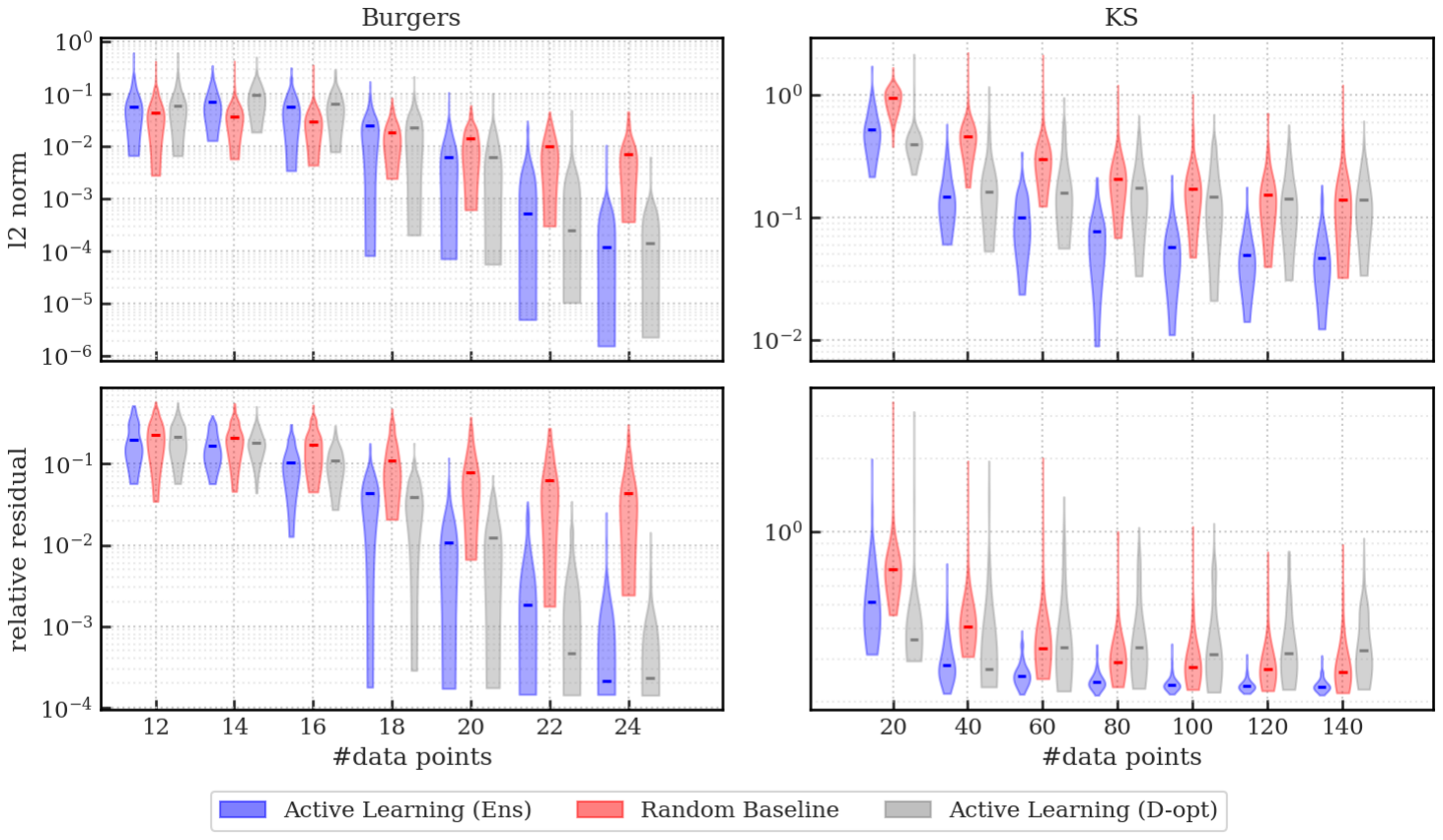}
        \vspace*{-2.4cm}
        \caption{Evaluation of the $\ell_2$ norm (top) and relative residual error (bottom) for the three sampling strategies: ensemble-based active learning (blue), random sampling (red), and D-optimal (grey). The relative residual quantifies how well the identified PDE fits held-out test data, computed as the mean absolute error between the library prediction $\boldsymbol{\Theta}\hat{\boldsymbol{\xi}}$ and the measured time derivative $\partial_t u$ over a fixed held-out set $\mathcal{T}$, normalized by the mean absolute value of $\partial_t u$ on the same set.}
        \label{fig:pderesults3}
        \vspace{-0.8cm}
    \end{center}
\end{figure}

For this work, this case is particularly interesting because the Burgers equation exhibits spatially localized dynamics, with sharp shock fronts concentrated in narrow regions of the domain. Away from these fronts, the solution varies smoothly and carries little information about the governing dynamics. Random sampling wastes a significant fraction of its budget in these uninformative smooth regions, while active learning concentrates samples where ensemble uncertainty is highest, corresponding to the shock front regions where the dynamically active structures appear (see Fig.~\ref{fig:pderesults-1}). Active learning thus offers a clear advantage over random sampling by efficiently targeting the shock fronts (see Fig~\ref{fig:pderesults3}).

\textbf{Kuramoto-Shivashinsky}. This equation is also a nonlinear \gls{pde} that originates in the study of flame front propagation~\cite{sivashinsky1977} and reaction-diffusion systems~\cite{kuramoto1978}, and has become a canonical benchmark for chaotic spatiotemporal dynamics:
\begin{equation} 
u_t + uu_x + u_{xx} + u_{xxxx} = 0, \quad x \in [0, L], \quad t > 0, \label{eq:ks} 
\end{equation} 
\noindent where the second-order term $u_{xx}$ destabilizes the system and the fourth-order term $u_{xxxx}$ provides dissipation, with the nonlinear term $uu_x$ transferring energy across scales. This balance produces spatiotemporal chaos for $L>44$~\cite{hyman1986}, making KS a widely used benchmark for chaotic systems.

The KS equation is spatially translation-invariant on a periodic domain, and once the solution settles onto the attractor, the ensemble uncertainty becomes approximately uniform across space. Despite this uniformity, actively sampled points consistently cluster in high-gradient regions (see blue dots in Fig.~\ref{fig:pderesults-1}), where the nonlinear interactions between the destabilizing and dissipative terms are most strongly expressed. This targeted sampling results in a lower equation discovery error than random sampling (see Fig.~\ref{fig:pderesults3}). Despite the visual similarity between the D-optimal and active learning sampling locations (see Fig.~\ref{fig:pderesults-1}), D-optimal performs closer to random sampling when evaluating the $\ell_2$ norm on the library matrix (see Fig.~\ref{fig:pderesults3}). This discrepancy highlights a fundamental difference between the two criteria: D-optimal maximizes the determinant of the information matrix $\Theta(\mathbf{X})^\top \Theta(\mathbf{X})$,
\begin{equation}
    \mathbf{x}^* = \underset{\mathbf{x} \in \mathcal{D}}{\mathrm{argmax}} \det\left(\Theta(\mathbf{X})^\top \Theta(\mathbf{X})\right),
    \label{eq:doptimal}
\end{equation}
a global geometric criterion that ensures spatial coverage but is agnostic to the local information content of the PDE solution. Active learning, by contrast, selects points where ensemble uncertainty is highest, directly targeting the most informative locations for characterizing the governing dynamics. Two sampling distributions can therefore appear spatially similar yet differ substantially in the quality of individual sample locations, with active learning consistently placing points at the most informative positions within the high-uncertainty interfaces rather than approximating them through a geometric proxy.

\begin{figure}[t!]
    \begin{center}
        \includegraphics[width=0.9\textwidth]{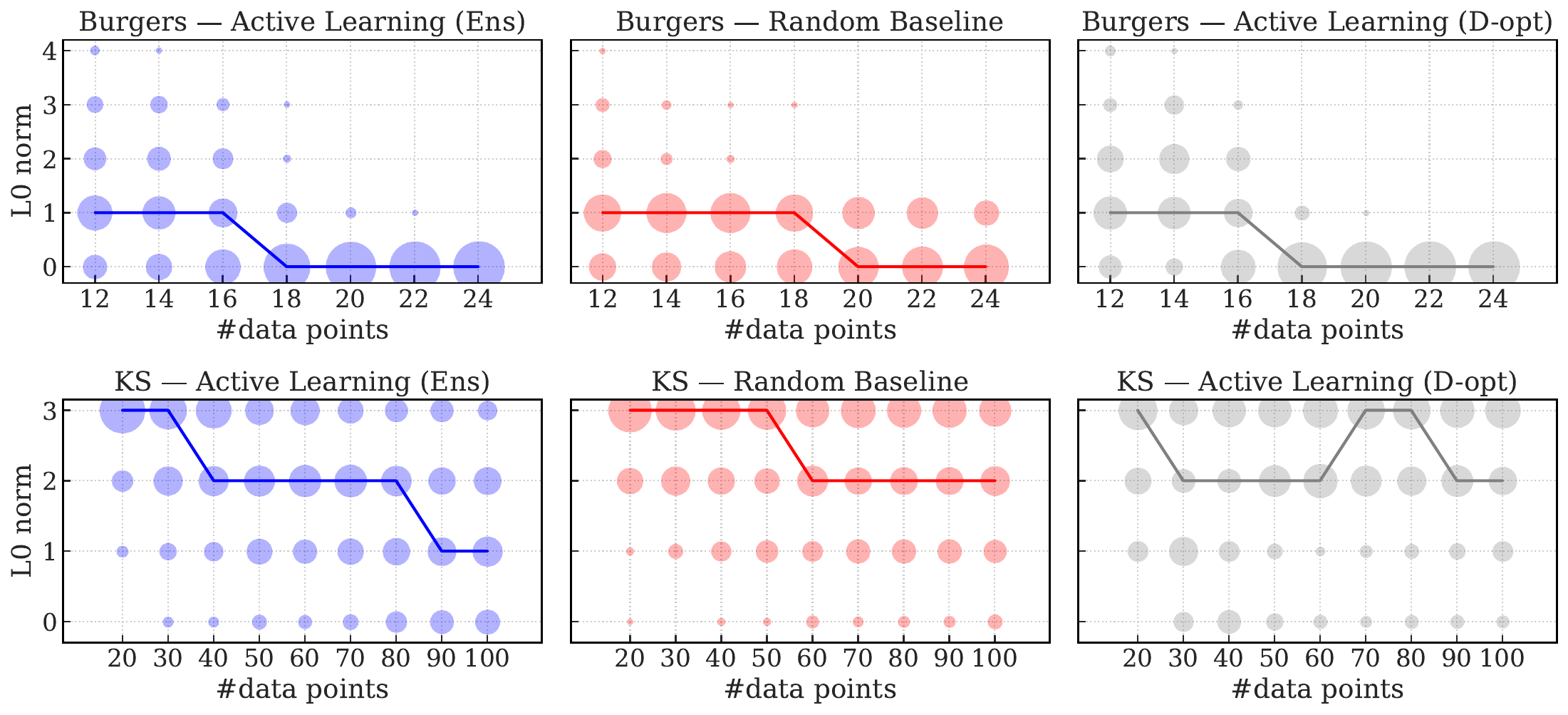}
        \vspace{-0.2cm}
        \caption{$\ell_0$ norm for Burgers and KS across the three sampling strategies: ensemble-based active learning (blue, left), random sampling (red, center), and D-optimal (grey, right). The top and bottom rows show results for the Burgers and Kuramoto--Sivashinsky equations, respectively. The data budget is shown up to the minimum number of points at which convergence was achieved.}
        \label{fig:pderesult2}
        \vspace{-1cm}
    \end{center}
\end{figure}

Finally, Fig.~\ref{fig:pderesult2} shows a comparison of the $\ell_0$ norm of the identified coefficient vector for both Burgers and KS across the three sampling strategies as a function of the data budget, where the median number of identified terms is shown as a connected line and the bubble size reflects the frequency of each outcome across runs. For Burgers, active learning reaches the correct sparsity level with fewer data points than random sampling and D-optimal, which require a larger budget to consistently threshold spurious terms. Interestingly, $uu_{xx}$ is the most persistent spurious term across all methods, but is suppressed significantly faster by active learning than by random sampling. For KS, all methods recover the correct terms $u_{xx}$, $u_{xxxx}$, and $uu_x$ rapidly, but the identification proves considerably more challenging due to the larger library from which spurious terms must be thresholded. In this regime, active learning achieves a consistently lower $\ell_0$ norm compared to both random sampling and D-optimal, with the latter two struggling to reach the correct sparsity level even as the data budget increases, as the spurious terms $u$, $u_x$, and $u_{xxx}$ remain difficult to threshold.

\vspace{-0.5cm}
        
\section{Discussion}
\label{Discussion}
\vspace{-0.4cm}
The present work addresses a fundamental question in data-driven dynamics discovery: \textit{how can we learn the system's governing equations with the minimal amount of information?} Two active learning strategies are developed, both grounded in the fast and accurate estimation of epistemic uncertainty through an ensemble of SINDy models. This ensemble-based approach provides a computationally efficient alternative to Bayesian methods while retaining the sparse, interpretable structure of SINDy.

For ODEs, an exhaustive analysis of the Lorenz system demonstrates that sampling from outside the attractor is critical to minimizing the data required to capture chaotic dynamics, and that the proposed active learning loop reliably identifies the correct governing equations with fewer than 50 data points across varying noise levels. For PDEs, a spatiotemporal sampling strategy is developed that iteratively locates the most informative points in the domain without requiring prior knowledge of the solution. Applied to the Burgers and Kuramoto--Sivashinsky equations, the method successfully identifies the governing dynamics with fewer than 24 and 100 data points, respectively, operating consistently in the ultra-low data limit. 

It is also important to note that active learning is most beneficial when the domain contains a mixture of informative and non-informative regions. When the dynamics are uniformly characterized across the domain, it offers little advantage over random sampling. This is evident in our comparison between the Burgers and Kuramoto–Sivashinsky equations: Burgers develops localized shocks that concentrate informative samples near steep gradients, whereas the spatiotemporal chaos of Kuramoto–Sivashinsky spreads information more uniformly across the domain. The same logic suggests a natural extension beyond dynamics discovery: in problems such as optimal control, where one seeks inputs that drive the system toward a desired state, samples could be chosen to be informative about the dynamics while also concentrating near control trajectories that minimize the cost. Zolman et al.~\cite{Zolman2025} explored this idea, proposing strategies that either exploit high-uncertainty regions to guide exploration during learning, or avoid them to remain within regions of trusted dynamics identification.

Overall, the presented framework demonstrates that uncertainty-guided sampling is an effective, simple, and practical strategy for model discovery in complex, high-dimensional nonlinear systems, achieving accurate coefficient recovery and robust sparse identification at a lower data cost of conventional approaches.

\vspace{-0.5cm}
\section*{Acknowledgments}
\vspace{-0.5cm}
ALJ and SLB acknowledge support from the National Science Foundation AI Institute in Dynamic Systems (grant number 2112085) and The Boeing Company.


\vspace{-0.5cm}

\begin{thebibliography}{61}
\providecommand{\natexlab}[1]{#1}
\providecommand{\url}[1]{\texttt{#1}}
\expandafter\ifx\csname urlstyle\endcsname\relax
  \providecommand{\doi}[1]{doi: #1}\else
  \providecommand{\doi}{doi: \begingroup \urlstyle{rm}\Url}\fi

\bibitem[Brunton and Kutz(2019)]{Brunton_Kutz_2019}
Steven~L. Brunton and J.~Nathan Kutz.
\newblock \emph{Data-Driven Science and Engineering: Machine Learning,
  Dynamical Systems, and Control}.
\newblock Cambridge University Press, 2019.

\bibitem[Shumaylov et~al.(2025)Shumaylov, Zaika, Scholl, Kutyniok, Horesh, and
  Schönlieb]{shumaylov2025}
Zakhar Shumaylov, Peter Zaika, Philipp Scholl, Gitta Kutyniok, Lior Horesh, and
  Carola-Bibiane Schönlieb.
\newblock When is a system discoverable from data? discovery requires chaos,
  2025.
\newblock URL \url{https://arxiv.org/abs/2511.08860}.

\bibitem[Brunton et~al.(2016)Brunton, Proctor, and Kutz]{brunton2016}
Steven~L. Brunton, Joshua~L. Proctor, and J.~Nathan Kutz.
\newblock Discovering governing equations from data by sparse identification of
  nonlinear dynamical systems.
\newblock \emph{Proceedings of the National Academy of Sciences}, 113\penalty0
  (15):\penalty0 3932--3937, 2016.
\newblock \doi{10.1073/pnas.1517384113}.

\bibitem[Angluin and Laird(1988)]{Angluin1988}
Dana Angluin and Philip Laird.
\newblock Learning from noisy examples.
\newblock \emph{Machine Learning}, 2\penalty0 (4):\penalty0 343--370, 1988.
\newblock ISSN 1573-0565.
\newblock \doi{10.1007/BF00116829}.

\bibitem[Maddu et~al.(2022)Maddu, Cheeseman, Sbalzarini, and Müller]{Madu2022}
Suryanarayana Maddu, Bevan~L. Cheeseman, Ivo~F. Sbalzarini, and Christian~L.
  Müller.
\newblock Stability selection enables robust learning of differential equations
  from limited noisy data.
\newblock \emph{Proceedings of the Royal Society A: Mathematical, Physical and
  Engineering Sciences}, 478\penalty0 (2262):\penalty0 20210916, 06 2022.
\newblock ISSN 1364-5021.
\newblock \doi{10.1098/rspa.2021.0916}.

\bibitem[Drgo{\v{n}}a et~al.(2025)Drgo{\v{n}}a, Nghiem, Beckers, Fazlyab,
  Mallada, Jones, Vrabie, Brunton, and Findeisen]{Drgona2025}
Jan Drgo{\v{n}}a, Truong~X. Nghiem, Thomas Beckers, Mahyar Fazlyab, Enrique
  Mallada, Colin Jones, Draguna Vrabie, Steven~L. Brunton, and Rolf Findeisen.
\newblock Safe physics-informed machine learning for dynamics and control,
  2025.
\newblock URL \url{https://arxiv.org/abs/2504.12952}.

\bibitem[Champion et~al.(2019)Champion, Lusch, Kutz, and Brunton]{champion2019}
Kathleen Champion, Bethany Lusch, J.~Nathan Kutz, and Steven~L. Brunton.
\newblock Data-driven discovery of coordinates and governing equations.
\newblock \emph{Proceedings of the National Academy of Sciences}, 116\penalty0
  (45):\penalty0 22445--22451, 2019.
\newblock \doi{10.1073/pnas.1906995116}.

\bibitem[Wagenmaker(2024)]{Wagenmaker2024}
Andrew Wagenmaker.
\newblock \emph{A Theory of Active Learning in Dynamic Environments}.
\newblock Ph.d.\ thesis, University of Washington, 2024.
\newblock URL
  \url{https://digital.lib.washington.edu/researchworks/items/8800ffc0-a15a-4817-aa05-2c173d1f098c/full}.
\newblock University of Washington ResearchWorks.

\bibitem[Zhao and Li(2022)]{zhao22a}
Zichen Zhao and Qianxiao Li.
\newblock Adaptive sampling methods for learning dynamical systems.
\newblock In \emph{Proceedings of Mathematical and Scientific Machine
  Learning}, volume 190 of \emph{Proceedings of Machine Learning Research},
  pages 335--350. PMLR, 2022.

\bibitem[Xu et~al.(2023)Xu, Ji, Li, and Lu]{Xu2023}
Pengcheng Xu, Xiaobo Ji, Minjie Li, and Wencong Lu.
\newblock Small data machine learning in materials science.
\newblock \emph{npj Computational Materials}, 9\penalty0 (1):\penalty0 42,
  2023.
\newblock ISSN 2057-3960.
\newblock \doi{10.1038/s41524-023-01000-z}.

\bibitem[Pickering et~al.(2022)Pickering, Guth, Karniadakis, and
  Sapsis]{Pickering2022}
Ethan Pickering, Stephen Guth, George~Em Karniadakis, and Themistoklis~P.
  Sapsis.
\newblock Discovering and forecasting extreme events via active learning in
  neural operators.
\newblock \emph{Nature Computational Science}, 2\penalty0 (12):\penalty0
  823--833, 2022.
\newblock \doi{10.1038/s43588-022-00376-0}.

\bibitem[Hu et~al.(2023)Hu, Camporeale, and Swiger]{camporeale2023}
A.~Hu, Enrico Camporeale, and B.~Swiger.
\newblock Multi-hour-ahead {Dst} index prediction using multi-fidelity boosted
  neural networks.
\newblock \emph{Space Weather}, 21\penalty0 (4):\penalty0 e2022SW003286, 2023.
\newblock \doi{10.1029/2022SW003286}.

\bibitem[Willard et~al.(2022)Willard, Jia, Xu, Steinbach, and
  Kumar]{Willard2022}
Jared Willard, Xiaowei Jia, Shaoming Xu, Michael Steinbach, and Vipin Kumar.
\newblock Integrating scientific knowledge with machine learning for
  engineering and environmental systems.
\newblock \emph{ACM Comput. Surv.}, 55\penalty0 (4), 2022.
\newblock ISSN 0360-0300.
\newblock \doi{10.1145/3514228}.

\bibitem[Geelen et~al.(2023)Geelen, Balzano, and Willcox]{Willcox2023}
Rudy Geelen, Laura Balzano, and Karen Willcox.
\newblock Learning latent representations in high-dimensional state spaces
  using polynomial manifold constructions, 2023.
\newblock URL \url{https://arxiv.org/abs/2306.13748}.

\bibitem[Della~Pia et~al.(2024)Della~Pia, Patsatzis, Russo, and
  Siettos]{Dellapia2024}
Alessandro Della~Pia, Dimitrios~G. Patsatzis, Lucia Russo, and Constantinos
  Siettos.
\newblock Learning the latent dynamics of fluid flows from high-fidelity
  numerical simulations using parsimonious diffusion maps.
\newblock \emph{Physics of Fluids}, 36\penalty0 (10):\penalty0 105187, 10 2024.
\newblock ISSN 1070-6631.
\newblock \doi{10.1063/5.0232378}.

\bibitem[Musekamp et~al.(2024)Musekamp, Kalimuthu, Holzm{\"u}ller, Takamoto,
  and Niepert]{musekamp2024}
Daniel Musekamp, Marimuthu Kalimuthu, David Holzm{\"u}ller, Makoto Takamoto,
  and Mathias Niepert.
\newblock Active learning for neural {PDE} solvers, 2024.
\newblock URL \url{https://arxiv.org/abs/2408.01536}.

\bibitem[Wu(2019)]{do2019}
Dongrui Wu.
\newblock Pool-based sequential active learning for regression.
\newblock \emph{IEEE Transactions on Neural Networks and Learning Systems},
  30\penalty0 (5):\penalty0 1348--1359, 2019.
\newblock \doi{10.1109/TNNLS.2018.2868649}.

\bibitem[Schaeffer et~al.(2020)Schaeffer, Tran, Ward, and Zhang]{Schaeffer2020}
Hayden Schaeffer, Giang Tran, Rachel Ward, and Linan Zhang.
\newblock Extracting structured dynamical systems using sparse optimization
  with very few samples.
\newblock \emph{Multiscale Modeling \& Simulation}, 18\penalty0 (4):\penalty0
  1435--1461, 2020.
\newblock \doi{10.1137/18M1194730}.

\bibitem[Seung et~al.(1992)Seung, Opper, and Sompolinsky]{seung1992}
Hyunjune~Sebastian Seung, Manfred Opper, and Haim Sompolinsky.
\newblock Query by committee.
\newblock In \emph{Proceedings of the Fifth Annual Workshop on Computational
  Learning Theory}, pages 287--294. ACM, 1992.
\newblock \doi{10.1145/130385.130417}.

\bibitem[Fasel et~al.(2022)Fasel, Kutz, Brunton, and Brunton]{Fasel2022}
Urban Fasel, J.~Nathan Kutz, Bingni~W. Brunton, and Steven~L. Brunton.
\newblock Ensemble-{SINDy}: Robust sparse model discovery in the low-data,
  high-noise limit, with active learning and control.
\newblock \emph{Proceedings of the Royal Society A}, 478\penalty0
  (2260):\penalty0 20210904, 2022.
\newblock \doi{10.1098/rspa.2021.0904}.

\bibitem[Fisher(1937)]{fisher1937design}
Ronald~A. Fisher.
\newblock \emph{The Design of Experiments}.
\newblock Oliver and Boyd, 1937.
\newblock URL \url{https://books.google.com/books?id=IMrtAAAAMAAJ}.

\bibitem[Atkinson and Donev(1992)]{atkinson1992}
Anthony~Curtis Atkinson and Alexander~Nikolaev Donev.
\newblock \emph{Optimum Experimental Designs}.
\newblock Oxford University Press, 08 1992.
\newblock ISBN 9780198522546.
\newblock \doi{10.1093/oso/9780198522546.001.0001}.

\bibitem[Kiefer and Wolfowitz(1959)]{kiefer1959}
Jack Kiefer and Jacob Wolfowitz.
\newblock Optimum designs in regression problems.
\newblock \emph{The annals of mathematical statistics}, 30\penalty0
  (2):\penalty0 271--294, 1959.

\bibitem[Atkinson et~al.(2007)Atkinson, Donev, and Tobias]{Atkinson2007}
Anthony~C. Atkinson, Alexander~Nikolaev Donev, and Randall~D. Tobias.
\newblock \emph{Optimum Experimental Designs, with {SAS}}.
\newblock Oxford University Press, 05 2007.
\newblock ISBN 9780199296590.
\newblock \doi{10.1093/oso/9780199296590.001.0001}.

\bibitem[Box(1992)]{box1992sequential}
George E.~P. Box.
\newblock \emph{Sequential Experimentation and Sequential Assembly of Designs}.
\newblock Report (University of Wisconsin--Madison. Center for Quality and
  Productivity Improvement). Center for Quality and Productivity Improvement,
  University of Wisconsin-Madison, 1992.
\newblock URL \url{https://books.google.com/books?id=UWnDNAAACAAJ}.

\bibitem[Xu et~al.(2024)Xu, Li, Bi, and Moeckel]{Xu2024}
Xukuan Xu, Donghui Li, Jinghou Bi, and Michael Moeckel.
\newblock Automl based workflow for design of experiments (doe) selection:
  benchmarking data acquisition strategies with simulation models.
\newblock \emph{Scientific Reports}, 14\penalty0 (1):\penalty0 32170, 2024.
\newblock \doi{10.1038/s41598-024-83581-3}.

\bibitem[Settles(2012)]{settles2012}
Burr Settles.
\newblock \emph{Active Learning}.
\newblock Synthesis Lectures on Artificial Intelligence and Machine Learning.
  Springer, Cham, 2012.
\newblock \doi{10.1007/978-3-031-01560-1}.

\bibitem[Holzmüller et~al.(2023)Holzmüller, Zaverkin, Kästner, and
  Steinwart]{holzmuller2023}
David Holzmüller, Viktor Zaverkin, Johannes Kästner, and Ingo Steinwart.
\newblock A framework and benchmark for deep batch active learning for
  regression, 2023.
\newblock URL \url{https://arxiv.org/abs/2203.09410}.

\bibitem[Fu et~al.(2013)Fu, Zhu, and Li]{Fu2013}
Yifan Fu, Xingquan Zhu, and Bin Li.
\newblock A survey on instance selection for active learning.
\newblock \emph{Knowledge and Information Systems}, 35\penalty0 (2):\penalty0
  249--283, 2013.
\newblock \doi{10.1007/s10115-012-0507-8}.

\bibitem[Adcock et~al.(2023)Adcock, Cardenas, and Dexter]{adcock2023}
Ben Adcock, Juan~M. Cardenas, and Nick Dexter.
\newblock {CS4ML}: A general framework for active learning with arbitrary data
  based on christoffel functions, 2023.
\newblock URL \url{https://arxiv.org/abs/2306.00945}.

\bibitem[Mania et~al.(2022)Mania, Jordan, and Recht]{mania2022}
Horia Mania, Michael~I Jordan, and Benjamin Recht.
\newblock Active learning for nonlinear system identification with guarantees.
\newblock \emph{Journal of Machine Learning Research}, 23\penalty0 (1), 2022.

\bibitem[Shields et~al.(2023)Shields, Gurley, Catarelli, Chauhan, Ojeda-Tuz,
  and Masters]{shields2023}
Michael~D. Shields, Kurtis Gurley, Ryan Catarelli, Mohit Chauhan, Mariel
  Ojeda-Tuz, and Forrest~J. Masters.
\newblock Active learning applied to automated physical systems increases the
  rate of discovery.
\newblock \emph{Scientific Reports}, 13\penalty0 (1):\penalty0 8402, 2023.
\newblock \doi{10.1038/s41598-023-35257-7}.

\bibitem[Lundby et~al.(2023)Lundby, Rasheed, Halvorsen, Reinhardt, Gros, and
  Gravdahl]{Elend2023}
Erlend Torje~Berg Lundby, Adil Rasheed, Ivar~Johan Halvorsen, Dirk Reinhardt,
  Sebastien Gros, and Jan~Tommy Gravdahl.
\newblock Deep active learning for nonlinear system identification, 2023.
\newblock URL \url{https://arxiv.org/abs/2302.12667}.

\bibitem[Delabays et~al.(2025)Delabays, De~Pasquale, D{\"o}rfler, and
  Zhang]{Delabays2025}
Robin Delabays, Giulia De~Pasquale, Florian D{\"o}rfler, and Yuanzhao Zhang.
\newblock Hypergraph reconstruction from dynamics.
\newblock \emph{Nature Communications}, 16\penalty0 (1):\penalty0 2691, 2025.
\newblock \doi{10.1038/s41467-025-57664-2}.
\newblock URL \url{https://doi.org/10.1038/s41467-025-57664-2}.

\bibitem[Horrocks and Bauch(2020)]{Horrocks2020}
Jonathan Horrocks and Chris~T. Bauch.
\newblock Algorithmic discovery of dynamic models from infectious disease data.
\newblock \emph{Scientific Reports}, 10\penalty0 (1):\penalty0 7061, 2020.
\newblock \doi{10.1038/s41598-020-63877-w}.
\newblock URL \url{https://doi.org/10.1038/s41598-020-63877-w}.

\bibitem[Kaiser et~al.(2018)Kaiser, Kutz, and Brunton]{Kaiser2018}
Eurika Kaiser, J.~Nathan Kutz, and Steven~L. Brunton.
\newblock Sparse identification of nonlinear dynamics for model predictive
  control in the low-data limit.
\newblock \emph{Proceedings of the Royal Society A: Mathematical, Physical and
  Engineering Sciences}, 474\penalty0 (2219):\penalty0 20180335, 11 2018.
\newblock ISSN 1364-5021.
\newblock \doi{10.1098/rspa.2018.0335}.

\bibitem[Zolman et~al.(2025)Zolman, Lagemann, Fasel, Kutz, and
  Brunton]{Zolman2025}
Nicholas Zolman, Christian Lagemann, Urban Fasel, J.~Nathan Kutz, and Steven~L.
  Brunton.
\newblock {SINDy-RL} for interpretable and efficient model-based reinforcement
  learning.
\newblock \emph{Nature Communications}, 16\penalty0 (1):\penalty0 10714, 2025.
\newblock \doi{10.1038/s41467-025-65738-4}.

\bibitem[Di~Fiore et~al.(2024)Di~Fiore, Nardelli, and Mainini]{DiFiore2024}
Francesco Di~Fiore, Michela Nardelli, and Laura Mainini.
\newblock Active learning and bayesian optimization: A unified perspective to
  learn with a goal.
\newblock \emph{Archives of Computational Methods in Engineering}, 31\penalty0
  (5):\penalty0 2985--3013, 2024.
\newblock \doi{10.1007/s11831-024-10064-z}.

\bibitem[Fan et~al.(2019)Fan, Jodin, Consi, Bonfiglio, Ma, Keyes, Karniadakis,
  and Triantafyllou]{Fan2019}
Dixia Fan, Gr{\'e}goire Jodin, Thomas~R. Consi, Luca Bonfiglio, Yue Ma, Lena~R.
  Keyes, George~E. Karniadakis, and Michael~S. Triantafyllou.
\newblock A robotic intelligent towing tank for learning complex
  fluid-structure dynamics.
\newblock \emph{Science Robotics}, 4\penalty0 (36):\penalty0 eaay5063, 2019.
\newblock \doi{10.1126/scirobotics.aay5063}.

\bibitem[Krogh and Vedelsby(1994)]{krogh1994}
Anders Krogh and Jesper Vedelsby.
\newblock Neural network ensembles, cross validation, and active learning.
\newblock In G.~Tesauro, D.~Touretzky, and T.~Leen, editors, \emph{Advances in
  Neural Information Processing Systems}, volume~7. MIT Press, 1994.

\bibitem[Vovk et~al.(2005)Vovk, Gammerman, and Shafer]{Vovk2005}
Vladimir Vovk, Alex Gammerman, and Glenn Shafer.
\newblock \emph{Algorithmic Learning in a Random World}.
\newblock Springer-Verlag, 2005.
\newblock ISBN 0387001522.

\bibitem[Fasel(2025)]{Fasel2025}
Urban Fasel.
\newblock Sparse identification of nonlinear dynamics with conformal
  prediction, 2025.
\newblock URL \url{https://arxiv.org/abs/2507.11739}.

\bibitem[Fung et~al.(2025)Fung, Fasel, and Juniper]{fung2025}
Lloyd Fung, Urban Fasel, and Matthew Juniper.
\newblock Rapid bayesian identification of sparse nonlinear dynamics from
  scarce and noisy data.
\newblock \emph{Proceedings of the Royal Society A: Mathematical, Physical and
  Engineering Sciences}, 481\penalty0 (2307):\penalty0 20240200, 2025.
\newblock ISSN 1364-5021.
\newblock \doi{10.1098/rspa.2024.0200}.

\bibitem[Rudy et~al.(2017)Rudy, Brunton, Proctor, and Kutz]{rudy2017}
Samuel~H. Rudy, Steven~L. Brunton, Joshua~L. Proctor, and J.~Nathan Kutz.
\newblock Data-driven discovery of partial differential equations.
\newblock \emph{Science Advances}, 3\penalty0 (4):\penalty0 e1602614, 2017.
\newblock \doi{10.1126/sciadv.1602614}.

\bibitem[Loiseau et~al.(2018)Loiseau, Noack, and
  Brunton]{Loiseau_Noack_Brunton_2018}
Jean-Christophe Loiseau, Bernd~R. Noack, and Steven~L. Brunton.
\newblock Sparse reduced-order modelling: sensor-based dynamics to full-state
  estimation.
\newblock \emph{Journal of Fluid Mechanics}, 844:\penalty0 459–490, 2018.
\newblock \doi{10.1017/jfm.2018.147}.

\bibitem[Messenger and Bortz(2021)]{messenger2021}
Daniel~A. Messenger and David~M. Bortz.
\newblock Weak sindy: Galerkin-based data-driven model selection.
\newblock \emph{Multiscale Modeling \& Simulation}, 19\penalty0 (3):\penalty0
  1474--1497, 2021.
\newblock \doi{10.1137/20M1343166}.

\bibitem[Reinbold et~al.(2020)Reinbold, Gurevich, and Grigoriev]{reinbold2020}
Patrick A.~K. Reinbold, Daniel~R. Gurevich, and Roman~O. Grigoriev.
\newblock Using noisy or incomplete data to discover models of spatiotemporal
  dynamics.
\newblock \emph{Phys. Rev. E}, 101:\penalty0 010203, 2020.
\newblock \doi{10.1103/PhysRevE.101.010203}.

\bibitem[Schaeffer and McCalla(2017)]{Schaeffer2017}
Hayden Schaeffer and Scott~G. McCalla.
\newblock Sparse model selection via integral terms.
\newblock \emph{Phys. Rev. E}, 96:\penalty0 023302, Aug 2017.
\newblock \doi{10.1103/PhysRevE.96.023302}.

\bibitem[Kaptanoglu et~al.(2021)Kaptanoglu, Callaham, Aravkin, Hansen, and
  Brunton]{alan2021}
Alan~A. Kaptanoglu, Jared~L. Callaham, Aleksandr Aravkin, Christopher~J.
  Hansen, and Steven~L. Brunton.
\newblock Promoting global stability in data-driven models of quadratic
  nonlinear dynamics.
\newblock \emph{Phys. Rev. Fluids}, 6:\penalty0 094401, 2021.
\newblock \doi{10.1103/PhysRevFluids.6.094401}.

\bibitem[Forootani et~al.(2023)Forootani, Goyal, and Benner]{ali2023}
Ali Forootani, Pawan Goyal, and Peter Benner.
\newblock A robust sindy approach by combining neural networks and an integral
  form, 2023.
\newblock URL \url{https://arxiv.org/abs/2309.07193}.

\bibitem[Melville and Mooney(2004)]{melville2004diverse}
Prem Melville and Raymond~J Mooney.
\newblock Diverse ensembles for active learning.
\newblock In \emph{Proceedings of the twenty-first international conference on
  Machine learning}, page~74, 2004.

\bibitem[Gao and Wang(2023)]{gao2023}
Wenhan Gao and Chunmei Wang.
\newblock Active learning based sampling for high-dimensional nonlinear partial
  differential equations.
\newblock \emph{Journal of Computational Physics}, 475:\penalty0 111848, 2023.
\newblock \doi{10.1016/j.jcp.2022.111848}.

\bibitem[Mohamad and Sapsis(2018)]{themis2018}
Mustafa~A. Mohamad and Themistoklis~P. Sapsis.
\newblock Sequential sampling strategy for extreme event statistics in
  nonlinear dynamical systems.
\newblock \emph{Proceedings of the National Academy of Sciences}, 115\penalty0
  (44):\penalty0 11138--11143, 2018.
\newblock \doi{10.1073/pnas.1813263115}.

\bibitem[Rudi et~al.(2018)Rudi, Calandriello, Carratino, and Rosasco]{Rudi2018}
Alessandro Rudi, Daniele Calandriello, Luigi Carratino, and Lorenzo Rosasco.
\newblock On fast leverage score sampling and optimal learning.
\newblock In S.~Bengio, H.~Wallach, H.~Larochelle, K.~Grauman, N.~Cesa-Bianchi,
  and R.~Garnett, editors, \emph{Advances in Neural Information Processing
  Systems}, volume~31. Curran Associates, Inc., 2018.

\bibitem[Pukelsheim(2006)]{Pukelsheim2006}
Friedrich Pukelsheim.
\newblock \emph{Optimal Design of Experiments}.
\newblock Society for Industrial and Applied Mathematics, 2006.
\newblock \doi{10.1137/1.9780898719109}.

\bibitem[Manohar et~al.(2018)Manohar, Brunton, Kutz, and Brunton]{Manohar2018}
Krithika Manohar, Bingni~W. Brunton, J.~Nathan Kutz, and Steven~L. Brunton.
\newblock Data-driven sparse sensor placement for reconstruction: Demonstrating
  the benefits of exploiting known patterns.
\newblock \emph{IEEE Control Systems Magazine}, 38\penalty0 (3):\penalty0
  63--86, 2018.
\newblock \doi{10.1109/MCS.2018.2810460}.

\bibitem[Burgers(1948)]{burgers1948}
Johannes~Martinus Burgers.
\newblock A mathematical model illustrating the theory of turbulence.
\newblock In Richard Von~Mises and Theodore Von~K{\'a}rm{\'a}n, editors,
  \emph{Advances in Applied Mechanics}, volume~1, pages 171--199. Elsevier,
  1948.
\newblock \doi{10.1016/S0065-2156(08)70100-5}.

\bibitem[Sivashinsky(1977)]{sivashinsky1977}
Grigory~I. Sivashinsky.
\newblock Nonlinear analysis of hydrodynamic instability in laminar
  flames---{I}. {D}erivation of basic equations.
\newblock \emph{Acta Astronautica}, 4:\penalty0 1177--1206, 1977.

\bibitem[Kuramoto(1978)]{kuramoto1978}
Yoshiki Kuramoto.
\newblock Diffusion-induced chaos in reaction systems.
\newblock \emph{Progress of Theoretical Physics Supplement}, 64:\penalty0
  346--367, 1978.

\bibitem[Hyman and Nicolaenko(1986)]{hyman1986}
James~M. Hyman and Basil Nicolaenko.
\newblock The {K}uramoto--{S}ivashinsky equation: A bridge between {PDE}s and
  dynamical systems.
\newblock \emph{Physica D}, 18:\penalty0 113--126, 1986.

\end{thebibliography}
\end{document}